\documentclass[10pt]{article} % For LaTeX2e
\usepackage[preprint]{tmlr}
\usepackage[utf8]{inputenc} % allow utf-8 input
\usepackage[T1]{fontenc}    % use 8-bit T1 fonts
\usepackage{hyperref}
\usepackage{url}
\usepackage{booktabs}       % professional-quality tables
\usepackage{amsfonts}       % blackboard math symbols
\usepackage{nicefrac}       % compact symbols for 1/2, etc.
\usepackage{microtype}      % microtypography
\usepackage{lipsum}
\usepackage{fancyhdr}       % header
\usepackage{graphicx}       % graphics

\usepackage{algorithm}
\usepackage{algpseudocode}
\usepackage{subfig}
\usepackage[export]{adjustbox}
\usepackage{enumitem}
\usepackage[normalem]{ulem}

\title{Assisting Human Decisions in Document Matching}

\author{\name Joon Sik Kim \email joonkim@cmu.edu \\
      \addr Carnegie Mellon University
      \AND
      \name Valerie Chen \email valeriechen@cmu.edu \\
      \addr Carnegie Mellon University
      \AND
      \name Danish Pruthi\thanks{Work done while at Carnegie Mellon University, prior to joining Amazon.} \email danish@hey.com\\
      \addr Amazon Web Services
      \AND 
      \name Nihar B. Shah \email nihars@andrew.cmu.edu \\
      \addr Carnegie Mellon University
      \AND
      \name Ameet Talwalkar \email talwalkar@cmu.edu \\
      \addr Carnegie Mellon University
}

\def\month{MM}  % Insert correct month for camera-ready version
\def\year{YYYY} % Insert correct year for camera-ready version
\def\openreview{\url{https://openreview.net/forum?id=XXXX}} % Insert correct link to OpenReview for camera-ready version

\begin{document}

\maketitle

%%% Abstract %%%
\begin{abstract}
    % Notwithstanding 
    % the remarkable 
    % improvement in performance of deep learning models, 
    % many applications warrant 
    
    % Decision makers are increasingly informed by machine learning models in many consequential applications. 
    % One such application is \emph{document matching}, which is applicable to scenarios 
    % ranging from academic paper-reviewer assignments
    % to screening of job applications, where decision makers identify relevant matches 
    % relying both on their expertise and outputs from machine learning models. 
    % Decision makers in these applications may desire additional information about the data or 
    % model outputs to make quicker and more accurate decisions. 
    % %However, there has been limited prior work studying what additional information can be helpful in such setting.
    % In this work, we devise a matching task on which different methods
    % extracting key information from the data and model 
    % are tested for their effectiveness in increasing the decision maker's accuracy and reducing time. 
    % A large-scale user study on the task
    % finds that some simple yet task-specific methods
    % could effectively benefit the decision makers, while more standard off-the-shelf methods 
    % like black-box model explanations, which are designed to be helpful by communicating the
    % reasoning behind the model output, may actually hinder them. 
    % We emphasize
    % understanding the decision makers' needs
    % for the task and 
    % designing solutions incorporating task-specific information,
    % rather than blindly relying on
    % model explanations.

    %% Shortened
    Many practical applications, ranging from 
    paper-reviewer assignment in peer review to 
    job-applicant matching for hiring,
    require human decision makers to identify relevant matches
    by combining their expertise with predictions from machine learning models. 
    In many such model-assisted document matching tasks, 
    the decision makers
    have stressed the need for assistive information 
    about the model outputs (or the data)
    to facilitate 
    their decisions. 
    In this paper,
    we devise a proxy matching task
    that allows us to 
    evaluate which 
    kinds of assistive information 
    improve decision makers' performance (in terms of accuracy and time).
    Through a crowdsourced ($N=271$ participants) study,
    we find that 
    providing black-box model explanations
    reduces users' accuracy on the matching task, 
    contrary to the commonly-held belief 
    that they can be helpful by allowing
    better understanding of the model.
    On the other hand, custom methods that are designed
    to closely attend to some task-specific desiderata 
    are found to be effective in improving user performance. 
    Surprisingly, we also find that the users' perceived 
    utility of assistive information
    is misaligned with their objective utility (measured through their task performance).

\end{abstract}

%%% Introduction %%%
\section{Introduction}
\label{sec:introduction}

\begin{figure*}[t]
    \centering
    \includegraphics[width=0.9\textwidth]{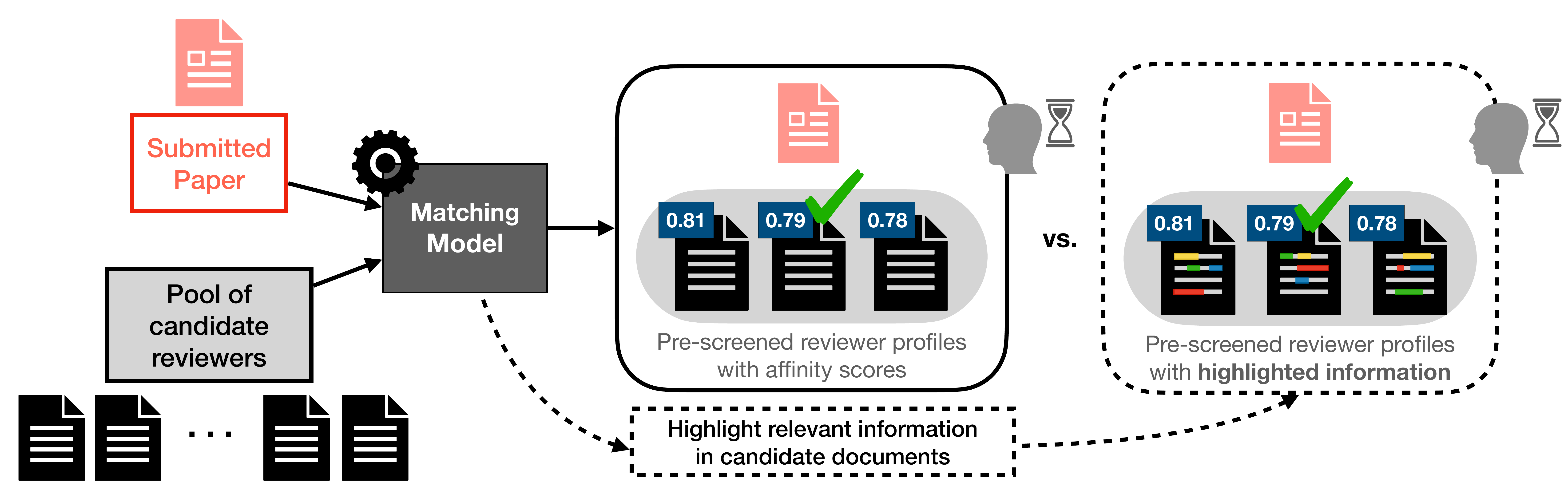}
    \caption{\small
    An example document matching application of peer review. For each submitted paper,
    % \ns{paper submission sounds like a verb. i think you are referring to a noun here. then use `submitted paper'. likewise elsewhere as well}\jkcomment{fixed}
    the matching model
    pre-screens a list of candidate reviewers via affinity scores (solid arrows). 
    Meta-reviewers, typically under a time constraint, then select the best match to the submitted paper among the pre-screened reviewer (box with a solid line). 
    We study whether providing additional assistive information, namely highlighting potentially relevant information in the candidate documents, can help the meta-reviewers make better decisions (dotted arrows and boxes). 
    We do so by focusing on a proxy matching task on a crowdsourcing platform 
    that is representative of real-world applications 
    not limited to peer review, 
    including recruitment and plagiarism detection 
    which follow the similar setup with different documents and decision makers. 
    % also follow this general setup 
    % involving different documents and different decision makers.
    % Because the affinity scores alone may be insufficient for the decision makers to base their decisions on, 
    % we study if highlighting additional relevant information 
    % with different methods can facilitate and improve their decisions (dotted arrows and boxes).
    }
    \label{fig:mainfig}
\end{figure*}

An important application 
in which human decision makers play a critical role, 
is
document matching,
% where 
% where human decision makers may benefit from AI assistance is in human-based document matching, 
i.e., when a \textit{query document} needs to be matched
to one of the many \textit{candidate documents} 
from a larger pool based on their relevance. 
Concrete instances of this setup include:
%  situations, including
\emph{academic peer review}, 
where meta-reviewers---associate editors in journals (e.g., \url{https://jmlr.org/tmlr/ae-guide.html}) or area chairs in conferences \citep{shah2022challenges} or program directors conducting proposal reviews \citep{kerzendorf2020distributed}---are asked to assign one or more candidate reviewers to submitted papers with relevant expertise based on their previously published work (illustrated in Figure~\ref{fig:mainfig}, solid arrows);
\emph{recruitment}, 
where recruiters screen through 
a list of resumes from candidate applicants 
for an available position at the company~\citep{schumann2020we, poovizhi2022}; and 
\emph{plagiarism detection}, 
where governing members (e.g., ethics board members of a conference, instructors of a course) 
review submissions
% relevant to submitted manuscripts 
to determine the degree of plagiarism~\citep{foltynek2019academic}.
Because the pool of candidate documents is typically large 
and the decision makers have limited time, 
they first use  
automated matching models 
to pre-screen the candidate documents. 
These matching models typically base their screening on
affinity scores, which measure the relevance of each candidate document to the query document~\citep{alzahrani2012, charlin2013toronto, cohan-etal-2020-specter, li2021algorithmic}. 
The human decision makers subsequently 
determine the best-matching document, 
taking both their expertise and the affinity scores 
computed by the matching models into account.
Such intervention by human decision makers is required for such tasks, 
as often times
either errors made by the models
are so consequential that they warrant human oversight,
or 
the overall performance
can be considerably improved by   
incorporating the domain knowledge of human experts.

Despite the 
growing prevalence of 
automated matching models and human decision makers 
working jointly for such practical matching tasks,
humans generally
find it difficult to completely rely 
on the models
due to a lack of assistive information other than the models' output itself. 
For instance, in peer review, 20\% of the meta-reviewers from past NLP conferences found the affinity scores from the matching model to be \emph{``not very useful or not useful at all''} in a recent survey~\citep{thorn-jakobsen-rogers-2022-factors}.
The survey also reports that the 
affinity scores rank the least important for the respondents, 
compared to more tangible and structured information about the candidate reviewers
such as whether they have worked on similar tasks, datasets, or methods. 
Additionally, the survey finds that providing just the affinity scores 
increases the meta-reviewers' workload 
as they \emph{``have to identify the information they
need from a glance at the reviewers’ publication
record.''} and \emph{``are presented with little structured information about the reviewers.''} 
Similarly, in hiring, the recruiters need to manually evaluate more profiles further
down the search result pages due to too generalized matches suggested by the model~\citep{li2021algorithmic}. 

% For instance, 20\% of the meta-reviewers from past NLP conference 
% who participated in a survey~\citep{thorn-jakobsen-rogers-2022-factors} 
% consider the affinity scores given to the reviewers by the matching model 
% \emph{``not very useful or not useful at all''} 
% in making their final 
% % \ns{this looks like decisions to accept or reject papers.. do you mean final assignments?}\jkcomment{fixed} 
% assignments.
% Furthermore, 
% the meta-reviewers in the survey indicate that they \emph{``are presented with little structured information about the reviewers.''} 
% In hiring, the recruiters need to manually evaluate more profiles further
% down the search result pages due to too generalized matches suggested by the model~\citep{li2021algorithmic}. 

% Such findings suggest that the decision makers experience 
% markedly increased workload in spite of the model's presence. 
% This finding suggests that meta-reviewers resort to perusing the actual content 
% of the submitted papers and the reviewers' past work, 
% increasing their workloads markedly. \vccomment{It would be nice to have more than one piece of evidence here.}
% \dd{FYI: I clipped the last phrase here as it felt redundant.}
% This is partly due to the  
%  information about the outputs of the matching model.
% with an increasing number of submissions and reviewers.
% the workload can quickly scale up.

% Task setup
% To better address such practical needs,
To address the lack of additional assistive information 
% other
% than the affinity scores 
in the document matching setup,
% \dd{"current setup ...  of document matching"},
we conduct the first evaluation of what additional information can
help the human decision makers to find matches \textit{accurately} and \textit{quickly} (Figure~\ref{fig:mainfig}, dotted arrows). 
To do so, we first design a proxy task of summary-article matching
that is representative of the general setup so that 
several methods providing different types of assistive information 
can be readily tested at scale via crowdsourced users (Section~\ref{sec:matching}). 
%\vccomment{do we describe what the proxy task is? i think that is missing.} 
% \jkcomment{i feel like describing the task is too much detail for the intro} 
% \vccomment{even a phrase like ``we design a summary-article proxy task'' would be more informative.}
The choice of proxy task 
addresses the logistical difficulty and expenses of directly experimenting with 
real domain-specific decision makers.
On this proxy task, we explore different classes of methods 
that have been previously suggested as tools for  users to
understand model outputs or  document content. 
% On this proxy task, we explore different classes of methods 
To standardize the format of assistance, 
we focus on methods that highlight 
assistive information within the candidate documents 
that the decision makers can utilize for matching (Section~\ref{sec:methods}):
\begin{itemize}
\item SHAP~\citep{lundberg2017unified}, a popular \textit{black-box model explanation}~\citep{doshi2017towards, chen2022interpretable},
highlights input tokens in the document 
that contribute both positively and negatively to the affinity scores. 
The utility of SHAP on several concrete downstream tasks remain controversial 
with conflicting 
% \ns{missing reference}\jkcomment{fixed}
results~\citep{kaur2020interpreting, jesus2021can, amarasinghe2022importance},
and has yet to be evaluated for its effectiveness in document matching.
\item BERTSum~\citep{liu-lapata-2019-text}, a state-of-the-art \textit{text summarization method}, which highlights key sentences 
in the candidate documents 
to help reduce the user's cognitive load for the task.
\item Two task-specific methods, that we design ourselves (Section~\ref{sec:methods}), to
% \ns{ (Section ??)}\jkcomment{fixed}
% that may be better suited for highlighting 
% details from the candidate documents relevant 
% to the details in the query -- a key condition to excel at the task.
% We design two methods to 
highlight details in the candidate documents relevant to the details in the query (by using sentence and phrase-level similarity measures).
% between the query and the candidate documents).
% \vccomment{note, avoid using candidate/candidates on their own because they refer to different things in different use cases}. 
% \vccomment{we should also say early, aka here, that we proposed these methods!}
% We evaluate two variants, 
\end{itemize}

With assistive information provided by these methods as treatments, 
and a control group provided with just the affinity scores
and no additional assistive information,
we conduct a pre-registered 
user study (with 271 participants) on a crowdsourcing platform.\footnote{Pre-registration document is available here: \url{https://aspredicted.org/LMM\_4K9}}
The study
% \ns{there seems to be something missing here?} \jkcomment{fixed}
% \dd{we have to be careful here, and qualify that our study \underline{on the proxy task of} ... reveals that: ... and later say that we believe these trends might be suggestive of other document matching applications.}
% large-scale
finds that (Section~\ref{sec:experiments}):
% where each participant is given a mixed set of easy and hard questions, whose difficulty is determined by whether the correct article can be easily selected based on the affinity scores alone (i.e., more work from the users is required in hard questions).
% We summarize our findings about the users' performance below, measured by accuracy and time.

\begin{itemize}
    \item Despite its usage
    % and projected successes 
    in numerous applications,
    SHAP decreases the participants' matching performance
    compared to the control group.
    %is expected to provide 
    %benefit over having no additional information (the control group).
    %Surprisingly, while it does help with the time, it does not help the users to be more accurate than the control, particularly for the hard questions. 
    % As the baseline accuracy of the matching model
    % on the hard questions (grey dotted line) is about the same, 
    % it may as well be more effective to rely on the model 
    % without human in the loop with SHAP.
    
    \item Contrary to the expectation that summarizing long articles could improve task efficiency, the summaries generated by BERTSum
    adversely impact the participants. Participants take longer to finish and are less accurate compared to the control group.
    % \dd{You can make "users" as the primary subject of this and the above bullet (rather than methods), and phrase those as "Users found SHAP to be ..." or "evidence suggests that users found summaries to be distracting"}
    % Just like SHAP, summarization did not provide edge 
    % over just relying on the matching model output as
    % seen from the baseline accuracy.
    
    \item Our task-specific methods,
    which are tailored to better identify details useful for the task,
    help the participants to be 
    quicker and more accurate compared to the control group. 
    % \vccomment{it would be better to say something about how our intuitions on what would help people on this use case turned out to be right.}
    % Their effectiveness is particularly prominent 
    % for hard questions, with better time and accuracy well beyond that of 
    % the control and the matching model itself.
    
    % For hard questions, the participants with information from the task-specific methods were significantly more accurate 
    % (59\% for semantic and 58\% for syntactic) 
    % compared to those without additional information (control, 42\%), 
    % followed by SHAP (32\%) and BERTSum (32\%).
    % For easy questions
    % the performance difference was not significant. 

    \item An overwhelming number of participants in \emph{all} treatment groups
    perceive that the highlighted information is helpful, 
    whereas the quantitative performance (accuracy and time) says otherwise.
    % The majority of the users in all treatment groups
    % replied positively when asked if 
    % the highlighted information was helpful, 
    % contradicting the matching performance 
    % measured by accuracy and time. 
    % \vccomment{This sentence could be better phrased.. smth like "The overwhelming response of users in all treatment groups was that highlighted information was helpful, despite the quantitative results on accuracy and time showing otherwise".}
    
\end{itemize}

The results suggest the benefits of designing 
task-specific assistive tools 
over general black-box solutions, 
and highlight the importance of 
quantitative evaluation of the methods' utility 
that is grounded on a specific task 
over subjective user perceptions~\citep{chen2022interpretable}.  
% TODO add below in the arxiv version
The code from the study is available at \url{https://github.com/wnstlr/document-matching}.
\section{Related Work}
\label{sec:related}

% Problems with how things are evaluated, how we address it
\noindent \textbf{Prior Evaluation of Assistive Information.} 
% \vccomment{need a summary sentence to talk about which assistive information we will cover? E.g., We survey the ways in which the types of assistive information considered in our study have been evaluated in prior work which include affinity scores, black-box model explanations, and text summarization techniques. Q: what is the goal of this subsection? is our goal to motivate what results may be? is our goal to say no one else has evaluated on this task before? this seems inconsistent between different paragraphs.}
We discuss how our proposed evaluation of different types of assistive information,
which include affinity scores, black-box model explanations, and text summaries, differs from how they have been previously evaluated. 
% \vccomment{it wasn't clear from the intro that the affinity score was one of the things we're evaluating}

% affinity scores
Affinity scores, computed by comparing the similarity of 
% using the \dd{by comparing the similarity of} 
representations learned by language models, are commonly used in practice to rank or filter the candidate
% \ns{some more papers onthis topic are cited in the 'practical concerns..' part but not here? let's keep both reference lists consistent since they are talking about the same thing}\jkcomment{fixed} 
documents~\citep{mimno07topicbased,rodriguez08coauthorsip, charlin2013toronto, tran17expertsuggestion,wieting2019simple, cohan-etal-2020-specter}
% ~\citep{dumais1992automating,mimno07topicbased,charlin2013toronto, cohan-etal-2020-specter, wu-etal-2020-mind}. 
Their quality has been 
evaluated both with or without human decision makers:
some may evaluate them based on the user's self-reported confidence score~\citep{mimno07topicbased},
while others may use performance from 
proxy tasks like document topic classification, where a higher test accuracy of the classification model using the learned representation indicates better ability to reflect more meaningful components in the documents~\citep{cohan-etal-2020-specter}.
%Their quality is often 
% \ns{not sure why this sentence is needed.. seems like a digression, plus misses out on the fact that some of these do evaluate baesd on reviewer-provided self-reported confidence} \jkcomment{modified to tell how they have been previously evaluated (both with or without human decision makers)}
% automatically evaluated on proxy tasks 
% like document topic classification or document citation prediction, with no involvement of human decision makers~\citep{cohan-etal-2020-specter}.
% \vcdelete{with automated approaches (i.e., without a human in the loop) 
% by using the learned representations as training features
% for proxy tasks like , where} 
% Using the learned representations as training features, a higher test accuracy of the trained classification model indicates that the representations capture more meaningful components from the documents. 
% Aside from such automated and user-independent evaluations, 
However, the utility of affinity scores for assisting human decision makers for the document matching task is less studied. 

% black-box model explanations (negative tone)
While information provided by black-box model explanations
have been evaluated for their utility to assist human decision makers
in various downstream tasks, 
the results have been lackluster. 
On the deception detection task, where users are asked to determine if a given hotel review is fake or not, prior work have shown that only some explanation methods 
improve a user's task performance~\citep{lai2019human, lai2020chicago}. 
\citet{Arora_Pruthi_Sadeh_Cohen_Lipton_Neubig_2022} further show that
none of the off-the-shelf explanations help 
the users better understand the model's decisions on the task. 
On more common NLP tasks like 
sentiment classification
and question-answering, 
providing explanations to the users
decreases the task performance compared to providing nothing 
when the model's prediction is incorrect~\citep{bansal2021does}.
For the fraud detection task with domain experts, 
providing some model explanations
showed conflicting effects on 
improving the performance~\citep{jesus2021can, amarasinghe2022importance}.
In this work, we expand user evaluations of 
black-box model explanations to the document matching task
and propose alternatives that could be more helpful.

Summaries generated by text summarization models~\citep{lewis-etal-2020-bart, liu-lapata-2019-text, shleifer2020pre, zhang2020pegasus} are typically either evaluated by metrics like ROUGE with respect to the annotated ground-truth summary
in a standardized dataset, 
or by a human's subjective rating of the quality.
To the best of our knowledge, 
the usefulness of
these automatically summarized information 
to the human decision makers
in concrete downstream tasks is rarely studied. 
Even for a few applied works that utilize these methods 
to practical documents 
in legal or business domains, the final evaluations
do not explore beyond 
these task-independent metrics~\citep{elnaggar2018multi, bansal2019review, huang2020legal}. 
% \vccomment{confused what "beyond these metrics" should imply to the reader, i.e., it's not clear to a reader why these metrics aren't good enough...}
In this work, we explicitly evaluate whether the generated summaries can help
improve the decision makers' task performance in document matching.
% While the use of these models beyond summarization tasks has been applied to 
% legal or business documents, 
% Although the use of these models beyond  summarization tasks has not been widely explored \dd{I don't think this claim is accurate, there likely are papers that explore applications of summarizing models}, these models may be able to assist document matching by reducing the amount of information the user needs to process. \dd{You can completely remove this paragraph, if you want.}

% \vccomment{i wonder if this section should go first.. first we motivate the problem, then we say how the methods we want to test have previously been evaluated / could connect to nihar's critique of "what do we expect" for each assistive info. 
% also i think we need a summary sentence here first too because it feels like 3 distinct paragraphs right now: There are a number of real-world document matching applications: peer review, hiring, and plagiarism detection. We discuss the needs for AI assistance in these applications.} 

\noindent \textbf{Practical Concerns in Document Matching Applications.} 
There are a number of real-world document matching applications including peer review, hiring, and plagiarism detection. 
For each application, we discuss practical issues that have been raised by users that can be mitigated by providing more assistive information about the data and the model.

In scientific peer review, submitted papers need to be matched to appropriate reviewers with proper expertise or experience in the paper's subject area.
First, a set of candidate reviewers are identified using an affinity scoring model based on representations learned by language models~\citep{charlin2013toronto, mimno07topicbased,rodriguez08coauthorsip,tran17expertsuggestion,wieting2019simple, cohan-etal-2020-specter}. Additional information such as reviewer bids or paper/reviewer subject areas may also be elicited~\citep{shah2017design, meir2020market, fiez2020super}. 
Based on this information, meta-reviewers may either be asked to directly assign one or more reviewers to each paper, or 
to modify the assignment that has been already made as they see appropriate.
% \jkdelete{the reviewers may be assigned in an automated fashion and meta reviewers are then asked to modify the assignment in a manner they think is appropriate.}
For example, in the journal Transactions on Machine Learning Research, for any submitted paper the meta-reviewer (action editor) is shown a list of all non-conflicted reviewers 
sorted according to the affinity scores.
% \jkdelete{along with their text-matching similarities. The list is ordered according to the text-matching similarities.}
The meta-reviewer may also click on any potential reviewer's name to see their website or list of publication. 
The meta-reviewer is then required to assign three reviewers to the paper based on this information. 
% \ns{wrote this para.. see what you think.. }
%This set of reviewers may then  be adjusted~\citep{shah2017design} based on other information like bidding information from candidate reviewers~\citep{thorn-jakobsen-rogers-2022-factors,meir2020market, fiez2020super}.  Meta-reviewers or Area Chairs (ACs), Finally, meta-reviewers or Area Chairs (ACs) finalize\ns{this doesn't sound right.. it reads like it is fully done by hand by meta reviewers} \jkcomment{isn't the finalization process handled by meta reviewers, whether they do it by hand or use some tools to make changes or just leave the automatically suggested results untouched?} the assignments by considering all available information.
% \vccomment{unsure what the latter part of this sentence is getting at}
% For some conferences, all the assignments are done manually. Larger conferences (e.g., machine learning conferences) use some combination of automated~\citep{goldsmith07aiconf,charlin2013toronto,stelmakh2018forall,kobren19localfairness,jecmen2020manipulation} and manual assignment methods. 
%\vccomment{the process described above seems to be the combined version, so it's weird to then immediately say that some assignments are done manually and expect the reader to follow how that works.}
However, a recent survey of meta-reviewers from past NLP conferences reveal
that the affinity scores alone are not as useful, and
most respondents prefer to see more tangible and structured information about the reviewers~\citep{thorn-jakobsen-rogers-2022-factors}.
% \jkcomment{The para that previously had a more detailed view on limitations of affinity scores in peer review is moved to the introduction (2nd para). Instead, here we braodly mention the main findings as above.}

% A recent survey of ACs from past NLP conferences found that roughly 20\% of respondents believed the affinity scores to be \emph{``not very useful or not useful at all''}~\citep{thorn-jakobsen-rogers-2022-factors}.
% The survey suggested that providing just the affinity scores increases the ACs' workload 
% as the ACs \emph{``have to identify the information they\ns{this is an extremely important paragraph motivating this work, and hence it should be there in the intro.. and in fact should be highlighted in the intro using bold/italics/quatation environment}\jkcomment{similar note is made in the introduction, first paragraph of page 2.}
% need from a glance at the reviewers’ publication
% record.''} 
% % because they needed to spend more time understanding 
% % the actual content of the papers and the reviewers' profiles. 
% % \vccomment{made some edits to shorten sentences and realized i dont actually understand these two lines.} 
% Additionally, the survey reported that the affinity scores rank the least important for the respondents, 
% compared to more tangible and structured information about the candidate reviewers
% such as whether they have worked on similar task, dataset, or methods.

In hiring, many companies resort to various algorithmic tools to efficiently filter and search for suitable candidates for a given job listing~\citep{fernandez2019ethical, black2020ai, poovizhi2022}. 
Many recruiters, while using these tools, express 
difficulties in reconciling a mismatch between algorithmic results and the recruiter's own assessments.
This is mainly attributed to ``too generalized and imprecise'' relevant matches 
suggested by the model, which lead to more ``manually evaluating more profiles further down the search result pages'' increasing the task completion time~\citep{li2021algorithmic}. 
Also, the general lack of understanding about the algorithmic assessments
makes the recruiters more reluctant to adopt them. 

In plagiarism detection, many existing software tools aim to
reduce the governing members' workload by
providing detailed information about the match, 
e.g., what specific parts of the query document are
identical or similar to parts of the candidate documents.
However, their performance in identifying
various forms of plagiarism
(e.g., ones involving paraphrasing or cross-language references) is still limited~\citep{jiffriya2021plagiarism}.
Also many existing tools lack user-friendly 
presentation of information 
that can better assist the task~\citep{foltynek2020testing}.
As the governing members need to ultimately assess the proposed evidence 
by the model to determine the degree of penalty~\citep{foltynek2019academic}, 
additional assistive information about the match 
may improve their experience.
\section{Task Setup and Methods}

In Section~\ref{sec:matching}, we describe the design
of a summary-article matching task, which is an instance of the document matching tasks.
We use this task as a proxy for other document matching tasks (e.g., matching reviewers to papers in peer-review), as it is 
more amenable for crowdsourcing experiments at scale. 
% of a proxy of the general document matching task called the summary-article matching task, that is suitable for use on a crowdsourcing platform at scale. 
% \dd{Maybe add a few lines about how running experiments on peer review is hard, and summary-article matching might serve as a good test bench for early prototyping and validation.}
The summary-article matching task addresses common 
difficulties encountered when directly experimenting on real-world applications like
recruiting real domain-specific decision makers (e.g., meta-reviewers in academia), 
building on complex systems in practice (e.g., internal systems that govern workflows in academic conferences), 
and coordinating logistical issues (e.g., longer turnaround for receiving feedback for each paper assignment).
Our task may also be useful
for 
early prototyping and validation of different methods.
 % \vcdelete{on which can be costly to iterate in production. }
Then in Section~\ref{sec:methods}, we present existing and our proposed methods that provide assistive information
that we evaluate with human users on the the summary-article matching task.

\subsection{Instantiating Document Matching}
\label{sec:matching}

% General task setup and why we design a proxy
% While it is ideal to set up document matching involving real decision makers and documents in applications like peer review and recruiting, there are limitations in doing so. 
% First, it is costly to recruit real decision makers for the study, which leads to inefficient turnarounds for observing which methods are promising or not.  
% Second, for real applications, not only the documents of interest are closed from public access, but also the ground-truth match within those documents (e.g. what match is considered good or bad) is not well-established. 
% To address these issues, we design a proxy task that resembles the real document matching task but leverages more accessible NLP datasets so that a general audience can be recruited and more methods can be tested efficiently. 
% In particular, 
% Instantiation with a specific dataset.
%% dataset used
%% how they were manually curated
%% characteristics of each problems, connecting back to the general task setup above. 

% \vccomment{this section is written too matter-of-factly and doesn't show the considerations made in trying to capture important aspects of the general task..first what are aspects of the general task (which include peer review, etc) that need to be captured?}

In the general document matching task, 
a matching model pre-selects a set of candidate documents
based on the affinity scores, which capture the relevance between the query document 
and the candidate document. 
% a user is given 
% a query document and a set of candidate documents,
% where the candidate documents are
% pre-selected by a matching model based on their relevance to the 
% query document. 
% This relevance is captured by affinity scores which
% are provided to the users. 
% \vccomment{i wonder if a better order is to say that the highest affinity scores are selected as the candidate documents, but it's not clear if the affinity score always finds the most relevant documents so that's why there is a user..}
These affinity scores facilitate filtering candidates
from a large pool of documents, but are nevertheless prone to errors.
The user therefore 
goes over the candidate documents with the scores 
and selects the most relevant candidate document. 
% For example, in the journal Transactions on Machine Learning Research, for any submitted paper the meta-reviewer (action editor) is shown a list of all non-conflicted reviewers 
% sorted according to the affinity scores.
% The meta-reviewer may also click on any potential reviewer's name to see their website or list of publication. 
% The meta-reviewer is then required to assign three reviewers to the paper based on this information.
A practical concern which we would like to address 
is when 
the affinity scores from the matching model alone
may not provide sufficient information 
to determine a match quickly and accurately.
% \vccomment{i dont think this point came across strongly enough in related work (or intro for that matter), also what does efficiently mean here..?}.
% in manually examining the data for a match efficiently.
% there is lack of useful information to utilize 
% to manually examine the data for a match efficiently,
% and the affinity scores from the matching model alone
% do not provide sufficient information. 
We outline how we instantiate the summary-article matching task that captures these details. 

\begin{figure*}[!t]
    \centering
    \includegraphics[width=\textwidth]{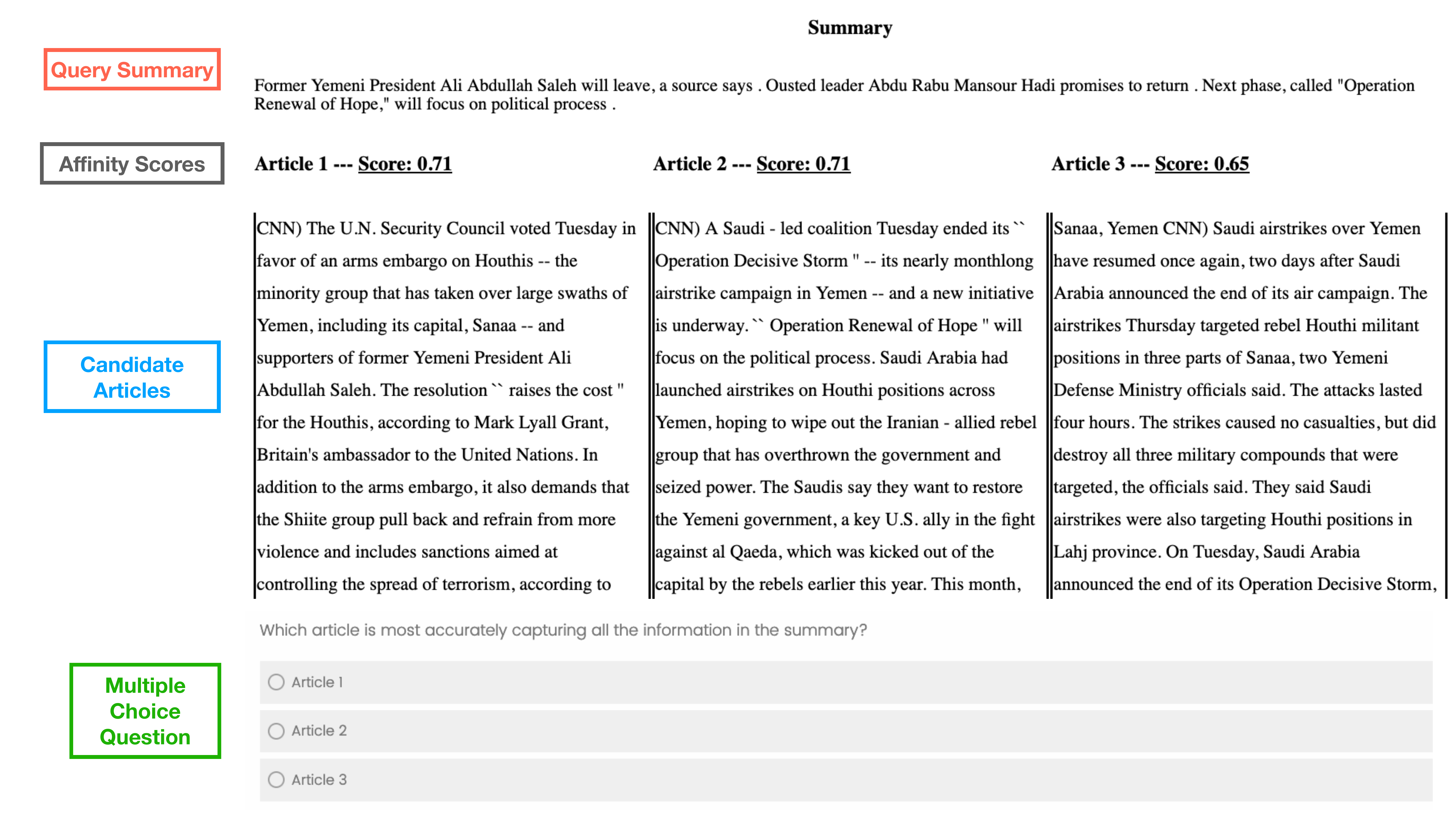}
    \caption{\small Interface for our summary-article matching task, an instance of the general document matching task. For each question, the participants are provided with the summary, three candidate articles to select from, and affinity scores for each candidate. The articles here are abridged to save space.}
    \label{fig:example-ui}
\end{figure*}

\noindent \textbf{Task setup.} We instantiate the general document matching tasks with a \textit{summary-article matching task}. Here, the query and candidate documents are each sampled from
human-written summaries and news articles 
in the CNN/DailyMail dataset~\citep{hermann2015teaching, see-etal-2017-get},
a common NLP dataset used for summarization task.
We select this dataset because the contents are accessible to a general audience, 
which enables us to evaluate a variety of assistive methods
by employing crowdworkers as in~\citet{lai2020chicago, wang2021explanations}.
So in our task, the participants are given a 
series of questions composed of
a query summary with three candidate articles\footnote{While the decision makers in a general matching task may observe more than three candidate articles, 
we devise a simpler instantiation here to reduce the complexity of the task, 
which will be better suited for the crowdsourcing task.},
and are asked to select an article from which the summary is generated under a time constraint (Figure~\ref{fig:example-ui}). 
% With this task, we address the aforementioned limitations by first making the query summary and the candidate articles to be more accessible to the general audience and those with basic reading comprehension skills can fully perform the task, and by secondly curating a set of articles so that a ground-truth match clearly exists. 
% The curation was performed in two steps, which we detail below: (1) filtering for similar articles (2) making manual adjustments for problem variations and ground-truths. 

As in the general document matching task, each candidate article is presented with an affinity score computed by a language model, which captures the similarity between the article and the summary.
The affinity scores are computed by taking a cosine similarity between the final hidden representation of a language model for the article and the summary~\citep{charlin2013toronto, wu-etal-2020-mind}. 
We use the representations from the DistilBART~\citep{shleifer2020pre} model fine-tuned on the CNN/DailyMail dataset.

\noindent \textbf{Question types.} 
In practice, there are some questions where the correct (document) match is obvious, 
whereas other questions require a more thorough inspection
of the specifics. 
For instance, in scientific peer review, 
a paper about a new optimization method in deep learning 
may be assigned to 
a broad range of candidate reviewers whose
general research area is within deep learning. 
However,
a reviewer who has worked both on 
optimization theory and deep learning
may be a better fit
compared to others who have primarily worked on
large-scale deep-learning based vision models. 
Even among the reviewers in optimization theory,
the reviewers who have worked on similar type of methods to the one proposed
may be better suited for the match. 
Such subtleties require more careful examination by the meta-reviewers. 
% \vccomment{this is a confusing first sentence and im not sure how the next sentence follows naturally..would suggest smth like "in practice, there are likely questions that are obvious which document is the best match for a query and questions that require more detailed inspection. We capture that via two types of questions.."}
% \vccomment{one more comment on this, i find this hard to relate to the general document matching task? mentioning via an example why hard questions are present in the real-world would be helpful. e.g., two candidate reviewers both work on X but one is a better fit because Y.}

We capture such scenarios by creating a data pool composed of two types of questions via manual inspection: 
easy and hard.  
Easy questions have candidate answers (articles) from different topics or events that are easily discernible from one another, and therefore can be easily matched correctly.
On the other hand, hard questions have candidate articles with a shared topic that only differ in small details, requiring a 
more careful inspection by the users. 
% \dd{I think we need to specify how we choose hard and easy questions?}

\begin{figure}[t]
    \centering
    \includegraphics[width=0.31\columnwidth]{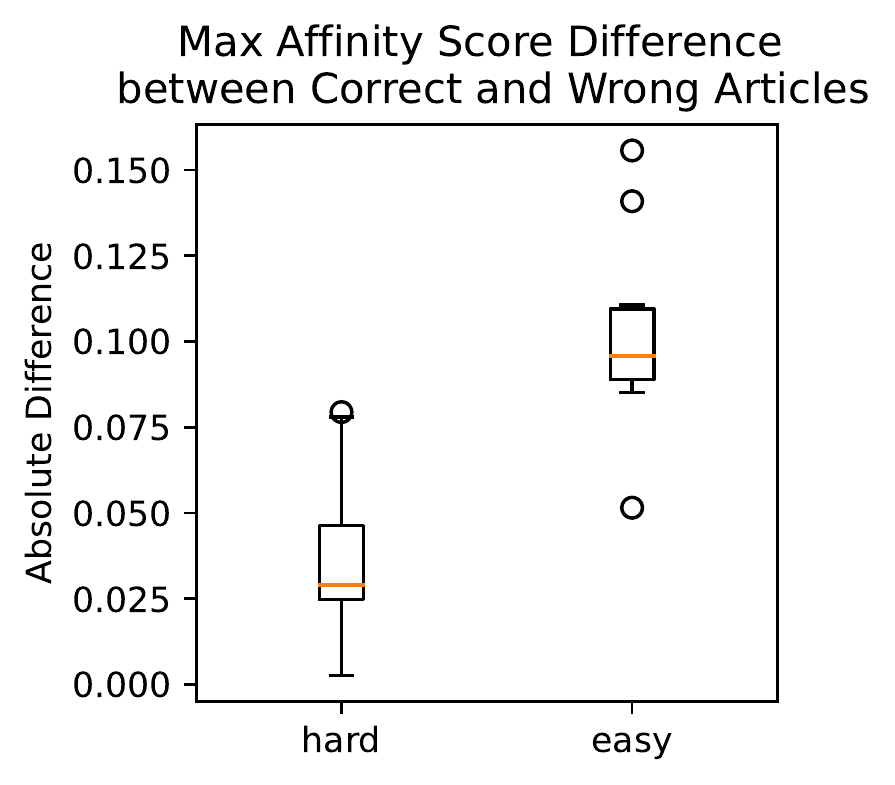}
    \caption{\small Distribution of affinity scores---computed by the matching model---for hard and easy questions. 
    % \vccomment{for someone who skims figure, what is the contet of this plot?} 
    The box plot shows maximum absolute difference in affinity scores between the correct and wrong candidate articles for each of the hard and easy questions. 
    % \dd{Since there are multiple wrong answers, does it compute the average difference over all the wrong answers?} \jkcomment{it is computing the max(wrong1, wrong2) for each question}
    The smaller the absolute difference is, the smaller the gap between the correct and the wrong article,
    making the scores less helpful in identifying the correct article (e.g., for the hard questions).
    % Similarly, the scores are accordingly expected to be less useful for the hard questions.
    % \vccomment{implying what..?}.
    }
    \label{fig:scorediff}
\end{figure}

\begin{figure*}[t]
    \centering
    \includegraphics[width=\textwidth]{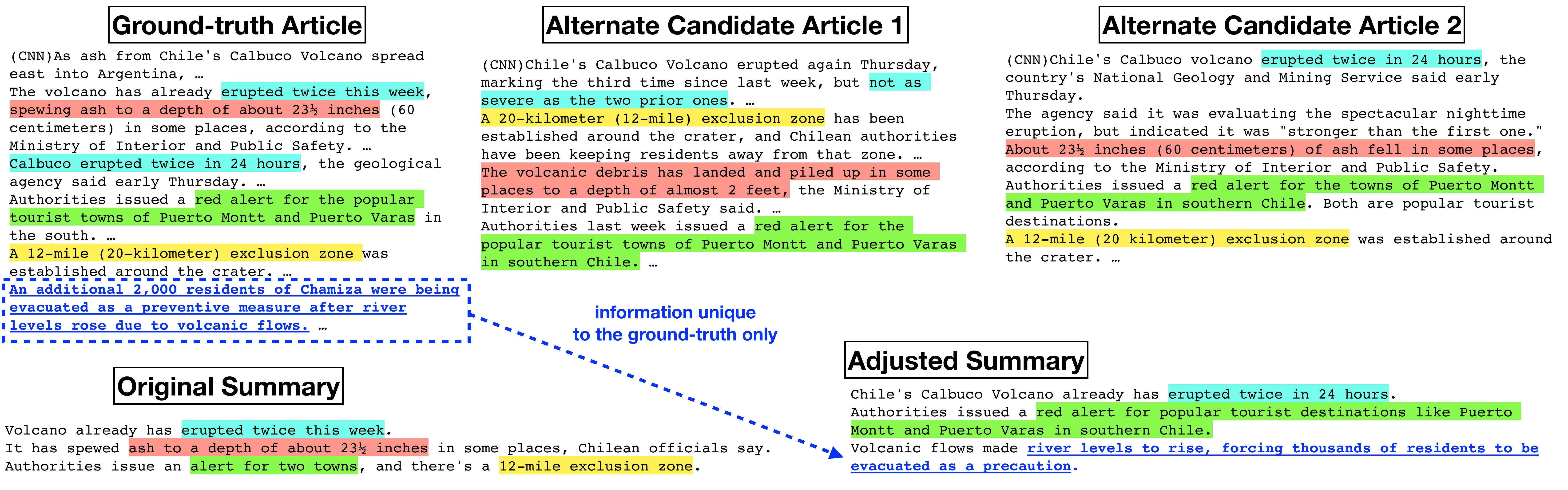}
    \caption{\small Ensuring a single correct match for the questions. 
    For an original summary sampled from the dataset, the three candidate articles appear to be all correct matches -- the critical information highlighted with different colors in the original summary is contained in all three candidate articles (highlighted with the same colors). 
    % An example of manual adjustments that were made to the query summary \vccomment{briefly mention what query summary is for someone who only skims the figures}. As highlighted in different colors, all important contents in the original summary from the dataset is already contained in not only the ground-truth article, but also other two candidate articles, making all three articles a valid correct answer \vccomment{this sentence is really long and confusing to read}. 
    To resolve having multiple correct matches for the question, 
    % To prevent this \vccomment{what does this refer to?}, 
    we manually extract information unique to the ground-truth article only (underlined text in dotted box) and add it to the adjusted summary to ensure that only the ground-truth article is the correct match for the summary.}
    \label{fig:manual-edit}
\end{figure*}

On easy versus hard questions, the affinity scores naturally show distinctive behaviors. 
The gap of the scores between the correct and the wrong matches is smaller for the hard questions than for the easy ones (Figure~\ref{fig:scorediff}). 
Because the scores for all candidate articles are similar to one another in the hard questions, the affinity scores
are not as helpful in identifying the best match.
Additionally, it is more likely that the candidate article
with the highest affinity score is not the correct match in the hard questions. 
If a hypothetical user was to simply select a candidate article 
which has the highest affinity score by completely relying on the matching model's output, 
they would be accurate only for 33.3 percent of the time 
for the hard questions,
compared to 100 percent of the time for the easy questions.
We believe that providing users with
assistive information 
might be critical for improving outcomes when making decisions on the hard questions,
when the model is less accurate 
and the correct match is more difficult to find. 

\noindent \textbf{Defining ground-truth matches.} A ground-truth match for a given summary and a set of candidate articles is necessary to measure participant performance.
To construct pairs of summary and candidate articles,
we first sample a summary-article pair from raw dataset and consider the article as the ground-truth for the given summary. 
We then select two other articles from the dataset which have
the highest affinity scores with respect to the given summary 
as the incorrect candidate articles for the given summary.

% Often times \dd{how often?}, 
There are several instances 
% \vccomment{how many or what percentage?}\jkcomment{uncertain; this was done manually to create a set of questions for the user study, so i am not sure about the precise value of “number of modified instances / number of data points checked”: the number of data points checked --- i did not keep track} 
where 
the two alternate candidate articles, which should be incorrect choices, 
may arguably be a suitable choice for the given summary. 
This happens because the dataset contains multiple articles covering the same event. 
To resolve this issue of having multiple ground-truths, 
we manually modify the given summary 
so that it 
% \vcdelete{better reflects the differences in details 
% among the candidate articles
% and}\vccomment{this is all vague, i think can leave out} 
is consistent only with the ground-truth article.
Specifically, we manually identify unique information in the ground-truth article that is not part of the alternate candidate articles and add that information to the summary (Figure~\ref{fig:manual-edit}).  

\subsection{Tested Methods}
\label{sec:methods}

% \jkcomment{method details and how they are presented to the users}
% \vcdelete{On the summary-article matching task described in Section~\ref{sec:matching},} 
In this section, we describe the methods used to highlight assistive information that we evaluate in our study
and how they are presented to the users.
% In this study, we consider several methods that highlight \vccomment{again, consistency in terminology} information in the 
% candidate news articles which can potentially help
% the participants make decisions more quickly and accurately in the summary-article matching task.
% We test four types of methods described below which cover multiple formats with distinct motivations \vccomment{the rest of this sentence is so vague}. 

% \vccomment{metacomment to discuss: a lot of this subsection overlaps with related work, we need to decide what should go where..}
% \jkcomment{shorten the moitvation part}

\textbf{Black-box Model Explanation.} Black-box model explanations include techniques that aim to highlight important input tokens for a model's prediction~\citep{simonyan2013deep, shrikumar2016not, sundararajan2017axiomatic, lundberg2017unified}. 
% It is notable that while the black-box model explanations have been actively tested on different downstream tasks with real users showing varying results~\cite{Arora_Pruthi_Sadeh_Cohen_Lipton_Neubig_2022, lai2019human, lai2020chicago, jesus2021can}, to the best of our knowledge, no results on the document matching task are known.
%In our setup, the model of interest is the affinity scoring model which computes how similar each candidate article is to the query summary. 
% As the presented affinity scores can carry insufficient information about the quality of the match, and overreliance on the scores can often lead to negative performance particularly for hard questions as described in Section~\ref{sec:matching}, we consider providing justification by pointing to the set of tokens in the article that contribute to the scores.
% Applied to our setup, these methods
% attribute scores to each input token from the candidate articles,
% indicating its importance to the predicted affinity scores by the matching model.
% These methods act on the affinity scoring model and return attributed scores for each token of the candidate articles, indicating its importance to the predicted affinity scores.
While there are several candidates to consider, we use a widely-applied method called SHAP~\citep{lundberg2017unified}.
SHAP assigns attribution scores to each input token
that indicate how much they contribute to the prediction output. 
% SHAP attributes the prediction output of the model 
% to the input features to satisfy
% various game-theoretic properties, 
% These attribution scores
% intuitively represent how much each input token contributes to
% the prediction output, either positively or negatively \vccomment{this is confusing.. can be described more concisely..}.
% \vcdelete{We found that SHAP may be a better candidate method to consider for the task
% compared to other prominent methods 
% like Integrated Gradients~\citep{sundararajan2017axiomatic} and
% Input x Gradients~\citep{shrikumar2016not},}
We select SHAP from a pool of prominent explanation methods 
(which include Integrated Gradients~\citep{sundararajan2017axiomatic} and
Input x Gradients~\citep{shrikumar2016not})
by examining how much the distribution of attribution scores
deviate from random distribution of attribution scores 
(see Appendix~\ref{appdx:posthoc} for details).

\begin{figure*}[t]
    \centering
    \includegraphics[width=\textwidth]{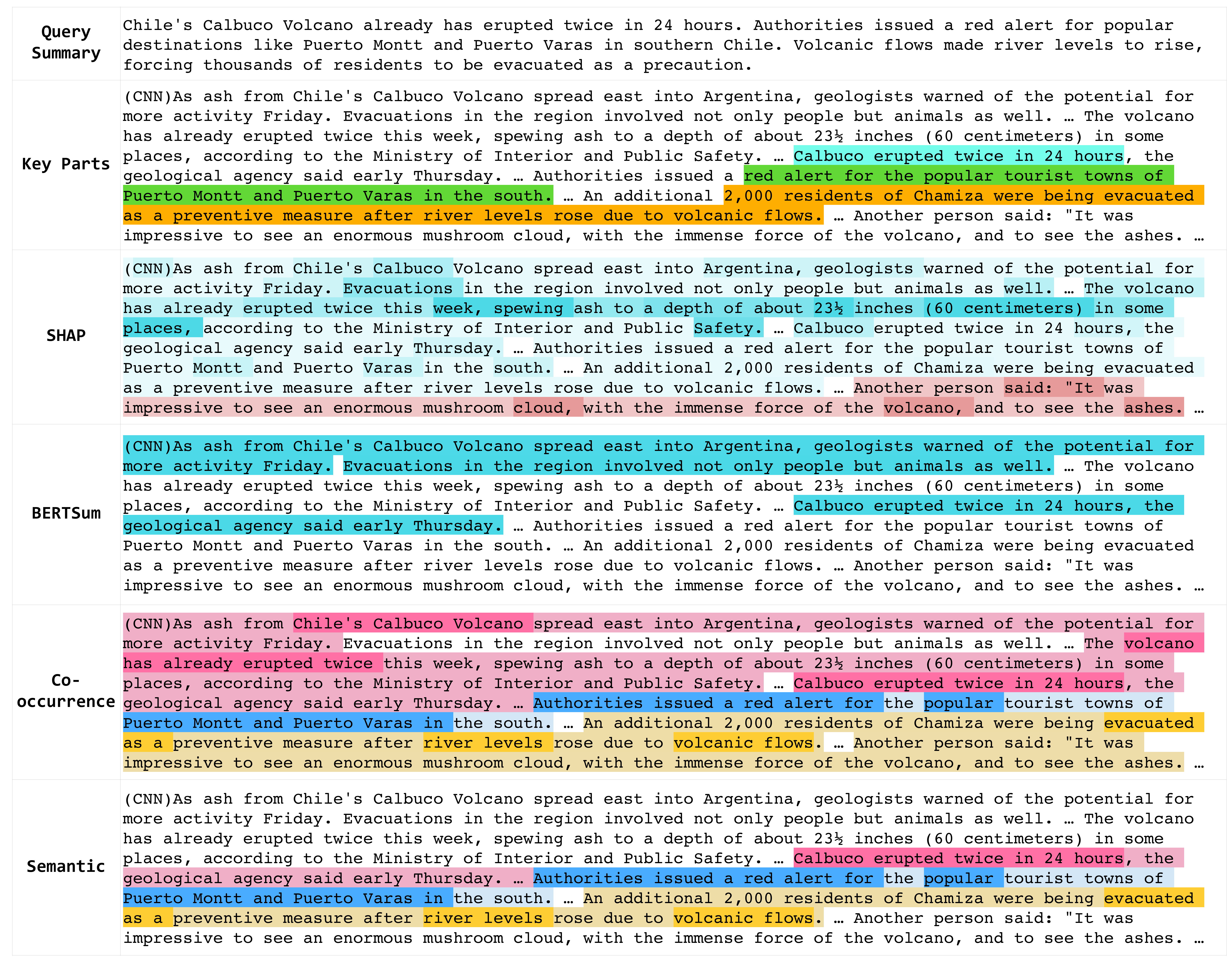}
    \caption{\small Highlighted information using different methods on the ground-truth article of the summary-article pair example in Figure~\ref{fig:manual-edit}. 
    Highlights for ``Key Parts'' (second row)
    indicate information relevant to the query summary (first row), 
    all of which ideally should be visibly highlighted by the methods that follow.
    SHAP (third row) and BERTSum (fourth row) fail to fully highlight all key parts. 
    Critically, they fail to visibly highlight the key part about river levels rising (yellow highlights in Key Parts), the unique information that distinguishes the ground-truth from other candidate articles (as described in Figure~\ref{fig:manual-edit}), which can directly impact the participant's performance. 
    On the other hand, our task-specific methods, both Co-occurrence (fifth row) and Semantic (sixth row) ones, are able to visibly highlight all key parts. 
    % While SHAP and BERTSum do not (or just slightly) highlight some of them \vccomment{for the reader who is skimming, it seems like SHAP/BertSum are highlighting a lot of things)}\jkcomment{i don't think bertsum is highlighting a lot of things? also i don't understand the point you are trying to make here}, Syntactic and Semantic methods \vccomment{say these are task-specific..}fully include them. Critically, failing to properly highlight the key part about river levels rising (yellow highlight in Key Parts), which is the unique information that distinguishes the ground-truth from other candidate articles (as described in Figure~\ref{fig:manual-edit}), can directly impact the participant's performance; SHAP and BERTSum fail to do so \vccomment{is this repetitive?}.
    % \vccomment{i find this caption hard to follow, it would be helpful to first overview in more precise words what is being shown and then get into how to interpret it.}
    }
    \label{fig:method-example}
\end{figure*}

We visualize SHAP (example shown in Figure~\ref{fig:method-example}, third row) by highlighting 
the input tokens
according to their attribution scores.
Tokens that contribute to increasing the affinity score (i.e., those with positive attribution scores)
are highlighted in cyan, while those that decrease the score (i.e., those with negative attribution scores)
are highlighted in pink.
The color gradients of the highlights indicate the magnitude of the attribution scores: the darker the color, the bigger the magnitude. 

\textbf{Extractive Summarization.}
% Explaining the decision of the matching model with black-box model explanations can help
% users understand the pre-selection process based on the affinity scores.
% However, there is additional information independent of the 
% matching model directly relevant to 
% what the users need to more efficiently complete the task. 
% \vcdelete{Because the users need to manually 
% process multiple articles at length
% in a short period of time in this task,
% they can benefit from 
% text summarization models}
Summarization methods are trained to select key information within a large body of text. 
These  methods can potentially help users process multiple lengthy articles in a shorter amount of time~\citep{liu-lapata-2019-text, zhong-etal-2020-extractive, lewis-etal-2020-bart, zhang2020pegasus}.
% For the summary-article matching task, they may be useful for the participants 
% to quickly grasp the content of each candidate articles given their length and limited time.
Summaries generated by these methods are typically either abstractive
(i.e., the summary is a newly-generated text that may not be part of the original text) 
or extractive (i.e., the summary is composed of text pieces extracted from the original text)~\citep{hahn2000}.
Because abstractive summaries are more susceptible to
hallucinating information not present in the original text~\citep{cao2018faithful, maynez-etal-2020-faithfulness, ji2022survey},
we focus on evaluating extractive summaries.
In particular, we use BERTSum~\citep{liu-lapata-2019-text}, 
which achieves state-of-the-art performance on the CNN/DailyMail dataset~\citep{pagnoni2021understanding},
to extract three key sentences from the article.
We visualize the extracted summary by highlighting the selected sentences from the original text with a single solid color (example shown in Figure~\ref{fig:method-example}, fourth row).

% Add a figure for each methods' output on sample text

\textbf{Task-specific Methods.} 
The summary-article matching task requires users to 
% \vccomment{i feel like the speed portion of this was not emphasized in the task set-up} 
accurately and quickly identify whether all details in the summary 
are correctly presented in each article.
% important, potentially low-level details in the article 
% are present in the low-level details in the summary \vccomment{confused why the summary has low-level details? it's a summary after all isn't it? maybe something like whether the article actually contains all of the details present in the summary?}. 
This is particularly challenging for hard questions, 
where the ground-truth can only be identified
by looking at the right part of the articles
due to their subtle differences.
% One potential drawback of earlier methods is that 
% \ddc{Earlier methods are not explicitly optimized for this particular objective:
% Model explanations are contingent on the matching model;
% it attempts to explain what the \emph{model} considers important,
% not necessarily what is important for \emph{human} users
% to perform well in the task. 
% General text summarization methods can be helpful for succinctly expressing 
% a high-level topic of the article
% but may lack the precision of picking the details directly related to the given summary.}{}
Next we propose two \textit{task-specific} methods that are more tailored to addressing this challenge.
% \dd{It is a bit ironical to outrightly reject the existing methods and then at the same time justify the need for conducting our study. I'd suggest that we later discuss in the discussion/conclusions why we believe the past methods fall short.}
% by working with the sentence-level information and pair-wise similarity, 
% the method is more attentive to low-level details that can better distinguish the candidates from one another. 

%%
The methods operate at sentence and phrase-level information 
in the summary and candidate articles.
Specifically,
we select and show the top $K$ sentences\footnote{We pick $K=3$, but this can be tuned for different levels of detail, depending on the length of the summary or the article.} from each article with
the highest similarity measure to each sentence in the summary. 
% \dd{Which sentence in the summary?}
To further provide more fine-grained detail on 
why that sentence could have been selected,
we then show exactly-matching phrases within those selected sentences. 
Essentially, the methods are designed to guide the users 
to relevant parts in the article for each summary sentence
by presenting the relevance hierarchically --  
by first showing the key sentences and then the key phrases within.
% \vccomment{some intuition why this addresses the above objective would be helpful here.}
% Consider a query summary $X_s = \{x_s^{(1)}, x_s^{(2)}, ..., x_s^{(n_s)}\}$ composed of $n_s$ sentences, and the candidate articles $X_d = \{X_{d_1}, X_{d_2}, X_{d_3}, ...\}$ where similarly each article $X_{d_i}$ is composed of $n_{d_i}$ sentences, $X_{d_i} = \{x_{d_i}^{(1)}, ...,x_{d_i}^{(n_{d_i})} \}$.
% As outlined in Algorithm~\ref{alg:taskspecific}, for each sentence in the query summary $X_s$, 
% we select top-$K$ sentences from each $X_{d_i}$, $i=1,2,3$, that are 
% the most relevant
% via some similarity measure {\sf sim}($\cdot, \cdot$). 
% We consider two versions of the method
% based on the type of similarity measure (line 5 of Algorithm~\ref{alg:taskspecific}) used to select the sentences: 
 
We consider two versions of the method which use different similarity measures to select the sentences:
% \dd{I strongly suggest NOT to name the first method as ``syntatic''. Alternative suggestions: ``N-gram Overlap''}

\begin{itemize}
\item \textbf{Co-occurrence method} uses F1 score of ROUGE-L~\citep{lin-2004-rouge}, a common performance metric used to capture the degree of n-gram co-occurrence between two texts. 
\item \textbf{Semantic method} uses the cosine similarity between the sentence representations from SentenceBERT~\citep{reimers-2019-sentence-bert}, a transformer model trained for sentence-level tasks. 
These scores are more sensitive to semantic similarities among texts like paraphrased components that may not be effectively captured by ROUGE-L.
\end{itemize}
% \vccomment{proposal to call these syntactic/semantic similarities instead of methods?}
% \jkcomment{using similiarities instead of methods can be confusing as semantic similarity itself already means something else (affinity scores are semantic similarities, and the semantic method uses this semantic similarity to filter and highilight)}

Once we select $K$ sentences based on the similarity measures,
we visualize them using
%different colors to indicate which sentence in the article is relevant to which sentence in the summary (
different colors to 
differentiate sentences in the article 
related to different sentence in the summary.
Like before, we use color gradients to indicate the magnitude of the similarity score for each sentence (the higher the similarity, the darker the color).
We then color the exactly-matching phrases using the darkest shade.
For instance, in Figure~\ref{fig:method-example} (fifth and sixth rows), the pink, blue and yellow highlights indicate relevant parts to first, second, and third sentence in the summary respectively. 
% The dark-colored phrases indicate the ones that appear verbatim in the query summary. 
% By presenting sentence and phrase-level information based on pair-wise similarity this way, 
% the methods are more attentive to low-level details \vccomment{would move this intuition earlier and set it up a little better (without relying on someone to understand the full method)}.
% 
We include additional examples from each of the explored methods in  Appendix~\ref{appdx:method-examples}. 

%%% Experiments %%%
\section{Experiments}
\label{sec:experiments}

We run a pre-registered\footnote{Pre-registration document is available at \url{https://aspredicted.org/LMM\_4K9}} user study on the summary-article matching task introduced in Section~\ref{sec:matching} to evaluate the methods described in Section~\ref{sec:methods} as treatment conditions. 
In this section, we outline the details of the user study (Section~\ref{sec:userstudy}), followed by our main hypotheses (Section~\ref{sec:hypothesis}) and results (Sections~\ref{sec:results}).  

\subsection{User Study Design}
\label{sec:userstudy}

% \ns{an annoying reviewer may complain that in this task we show only three options to the reviewer, whereas in the paper matching task they may be shown a sorted list of all reviewers (ranked by the affinities). we should maybe at least comment on it saying that this is a slightly simplified instantiation of that task .. perhaps as a footnote somwhere.. is it already written anywhere?} \jkcomment{added a footnote in page 5 Section 3.1 Task Setup, when we mention figure 2.}

We present $16$ questions to each participant. 
The 16 questions comprise 
$4$ easy and $12$ hard questions in random order. 
Participants complete all questions in one sitting. 
For each question, participants
see a query summary followed 
by three longer candidate articles (see Figure~\ref{fig:example-ui} for an example). 
To incorporate the time constraints 
typical decision makers may face 
in practical settings, as similarly done in \citep{pier2017your},
we limit participants to spend  $3$ minutes
% participants of the study have $3$ minutes
% \ns{There is precedence to putting a time limit in order to replicate time constraints in real world review systems. see~\cite{pier2017your} ``We timed each meeting to ensure that the panel spent approximately 15 minutes reviewing each application, creating a level of time pressure similar to what reviewers report experiencing in actual NIH study sections.''  consider citing it to motivate this part of the design} \jkcomment{cited}
to answer each question, 
after which they automatically see the next question. 
We offer bonus payments 
to encourage high-quality responses 
in terms of both accuracy and time (more details in Appendix~\ref{appdx:payments}). 
% \dd{We should include the payment, bonus details here.}
% \vccomment{i still think the time aspect isnt well motivated enough throughout}. 

We recruit $275$ participants from a balanced pool of adult males and females located in the U.S. with minimum approval ratings of $90\%$ on Prolific (\url{www.prolific.co}), with diverse demographic background (more details in Appendix~\ref{appdx:demoinfo}).
The sample size is determined from Monte Carlo power analysis based on data collected from a separate pilot study, for a statistical power above 0.8 (more details in Appendix~\ref{appdx:samplesize}).
Each participant is then randomly assigned to one of five groups: 
\begin{itemize}
    \item \emph{Control}: participants see the basic information (summary, articles, affinity scores)
    \item \emph{SHAP}: participants see the basic information + highlights from SHAP
    \item \emph{BERTSum}: participants see the basic information + highlights from BERTSum
    \item \emph{Co-occurrence}: participants see the basic information + highlights from Co-occurrence method
    \item \emph{Semantic}: participants see the basic information + highlights from Semantic method
\end{itemize} 
We include two attention check questions in the study in addition to the $16$ questions above.
$271$ out of $275$ participants pass both attention-check questions, and we exclude responses from the $4$ non-qualifying participants from our further analysis. 
We include mode details about the user study in Appendix~\ref{appdx:study-design}.

\subsection{Main Hypotheses}
\label{sec:hypothesis}

We pose the following null hypotheses with two-sided alternatives about the mean accuracy of participants on the hard questions, using different kinds of assistive information:
\begin{itemize}[label={}]
\item (\textbf{H1}) The mean accuracy of participants using SHAP is not different from that of the control. 
\item (\textbf{H2}) The mean accuracy of participants using BERTSum is not different from that of the control.
\item (\textbf{H3}) The mean accuracy of participants using  Co-occurrence method is not different from the control. 
\item (\textbf{H4}) The mean accuracy of participants using  Semantic method is not different from the control.
\end{itemize}

To test each hypothesis, we compare the mean accuracy of the participants in different treatment settings against the control group with two-tailed permutation tests, where the test statistic is the difference in the mean accuracy. 
We account for multiple comparisons with Sidak correction~\citep{sidak1967} for the family-wise error rate of 0.05. 
% \dd{I know Sidak correction is popular, but still might help to add a citation.}

\subsection{Results}
\label{sec:results}
We now discuss the participants' task accuracy (Section~\ref{sec:accuracy}), completion time (Section~\ref{sec:time}), 
and qualitative responses (Section~\ref{sec:qualresponse}) in different treatment groups.
% the participants
% differenc among different treatments, as well\vccomment{mention we look at accuracy, time, and qualitative?}. 

\begin{figure*}[t]
    \centering
    % \subfloat[Average accuracy with 95\% confidence intervals. \textbf{Bold} means statistically significant compared to control.]{
    %     \adjustbox{valign=c}{\begin{tabular}{c c c }
    %          & Hard & Easy \\ \hline \hline
    %         Control & 0.47 [0.42, 0.51] & 0.93 [0.90, 0.97] \\SHAP & \textbf{0.35 [0.32, 0.39]} & 0.87 [0.81, 0.92] \\BERTSum & \textbf{0.37 [0.32, 0.42]} & 0.92 [0.88, 0.97] \\Syntactic & \textbf{0.58 [0.53, 0.63]} & 0.94 [0.90, 0.98] \\Semantic & \textbf{0.59 [0.54, 0.64]} & 0.93 [0.89, 0.97]
    %     \end{tabular}
    %     \label{tab:accuracy}}}
    \subfloat[Average accuracy with 95\% confidence intervals.]{           
        \includegraphics[width=0.58\linewidth,valign=c]{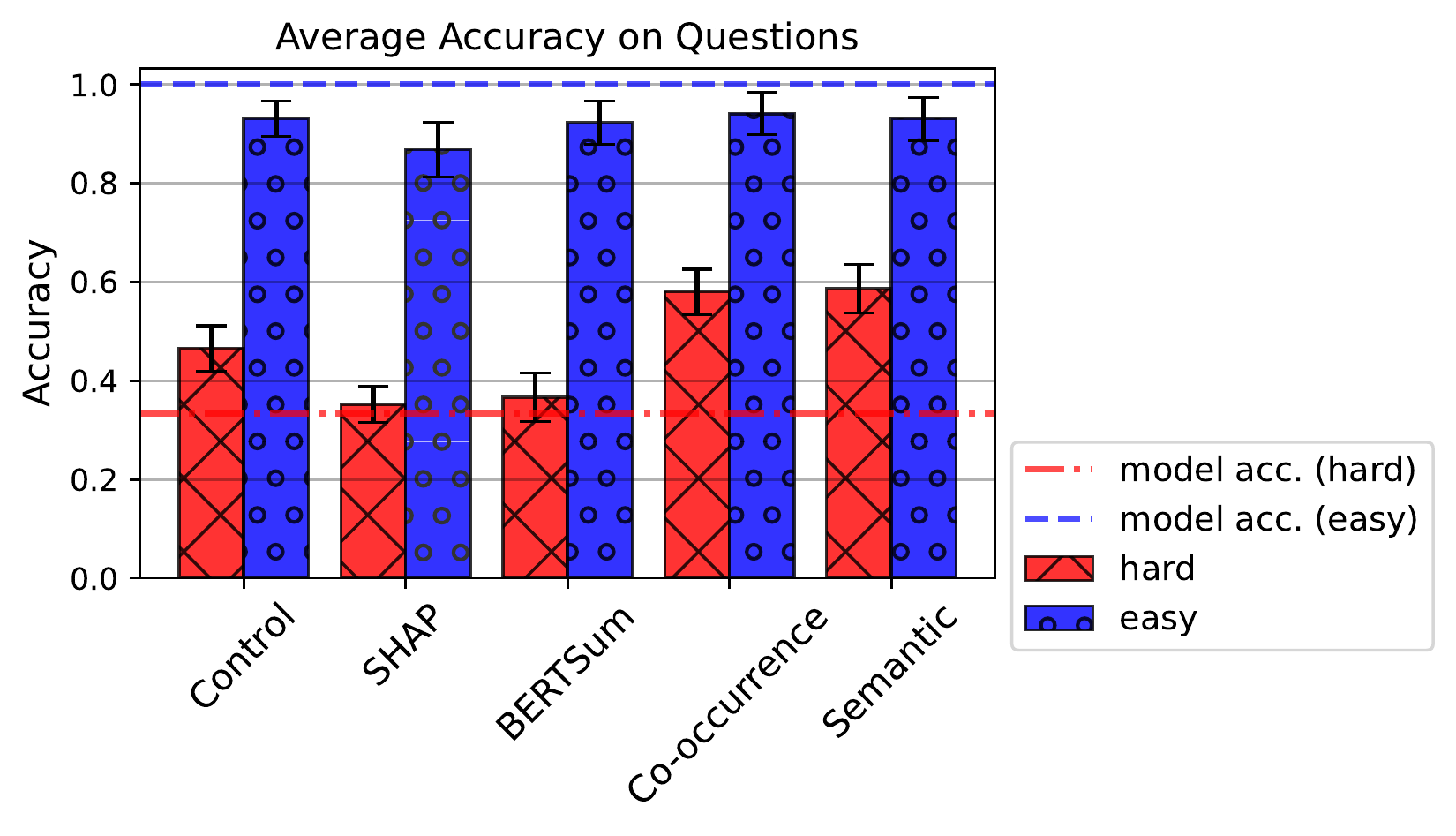} 
        \label{fig:accuracy}      
    }
    \subfloat[Average time with 95\% confidence intervals.]{
        \includegraphics[width=0.41\linewidth,valign=c]{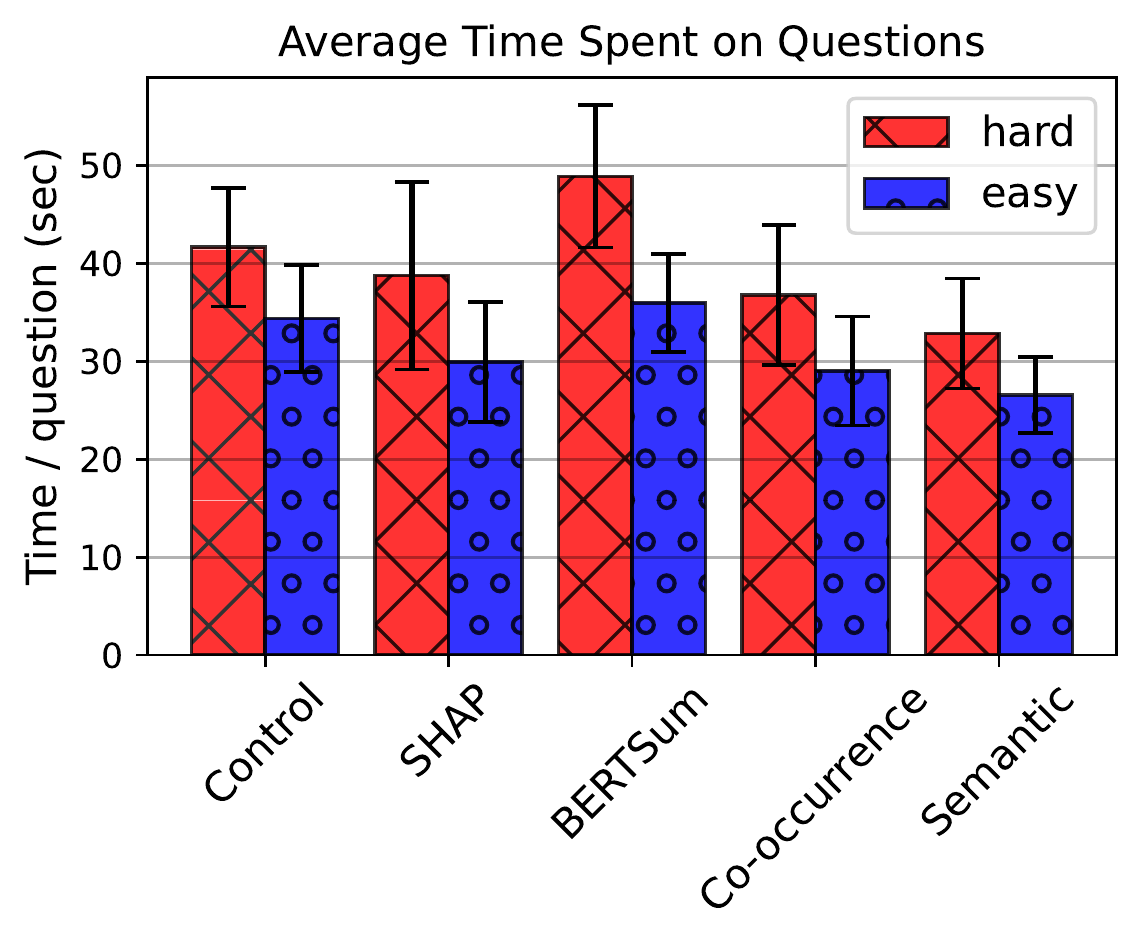}
        \label{fig:time}
    }
    \caption{\small
    (a) On hard questions, we observe significantly higher accuracy in groups presented with assistive information from Co-occurrence and Semantic methods compared to the control, and lower accuracy in groups presented with assistive information from SHAP and BERTSum. 
    The dotted lines indicate the accuracy of the matching model, i.e., 
    accuracy when selecting the article with the highest affinity score. 
    For easy questions, it is more effective to simply follow the affinity scores without
    support from additional assistive information. 
    % \vccomment{not clear that accuracy of matching model = follow affinity score}, \vccomment{say something about not needing to use the methods that were tested?} 
    For hard questions where the correct match is less obvious, using Co-occurrence or Semantic methods may be effective.  
    (b) We observe a lower average time for groups given SHAP, Co-occurrence, and Semantic methods compared to control in both types of questions, but higher average time for BERTSum.}
    \label{fig:accuracy-all}
\end{figure*}

\subsubsection{Accuracy Difference}
\label{sec:accuracy}

We find a statistically significant difference in the mean accuracy of all treatment groups compared to the control and reject the null hypotheses \textbf{H1} through \textbf{H4} in Section~\ref{sec:hypothesis}.
\begin{itemize}
    \item Participants using SHAP perform significantly \textit{worse} than the control ($p=0.001599 < 0.05$).
    \item Participants using BERTSum perform significantly \textit{worse} than the control ($p=0.0056 < 0.05$).
    \item Participants using Co-occurrence method perform significantly \textit{better} than the control ($p=0.002997 < 0.05$).
    \item Participants using using the Semantic method perform significantly \textit{better} than the control ($p=0.002997 < 0.05$).
\end{itemize}

% \dd{Might help to rephrase the above results with ``\underline{People with} SHAP perform signi...''}

% SHAP and BERTSum perform significantly \textit{worse} than the control with adjusted p-values of $p=0.001599$, $p=0.0056$ respectively (< $0.05$). 
% On the other hand, both Syntactic and Semantic methods are both significantly \textit{better} than the control with adjusted p-values of $p=0.002997$, 
%\ns{just to confirm this is not a typo and both p values are identical}
% $p=0.002997$ respectively (< $0.05$). 
% % Mean accuracy for each group and corresponding confidence intervals are presented in Table~\ref{tab:accuracy}, with statistically significant results in bold.
% These results corroborate \textbf{H4} and disproves all other hypotheses in Section~\ref{sec:hypothesis}: the Semantic method was shown to be useful as hypothesized, the Syntactic method was surprisingly useful, and SHAP and BERTSum impaired the accuracy of the users. 

Comparing the participants' accuracy against the model accuracy 
on different question types, 
we verify that the assistive information is particularly helpful
for the hard questions (Figure~\ref{fig:accuracy}).
Note that the model is only accurate around 33.3\% of the time in hard questions (red dotted line) while being 100\% accurate on easy questions (blue dotted line). 
The information from Semantic method is the most effective for the hard questions
with the highest average accuracy of 58.6\%, 
which is a 26\% increase in accuracy
compared to the control (46.6\%) and a 77\% increase compared to the model accuracy (33.3\%),
while SHAP (35.2\%) and BERTSum (36.7\%) remain less effective (red checker-patterned bars). 
On the other hand, there appears no significant difference in accuracy among the methods
for the easy questions (blue dotted bars), 
all of them slightly less accurate than the model accuracy. 
The results suggest that 
while it is more efficient to rely on the affinity scores
for the easy questions,
assistive information via Semantic methods can be 
particularly helpful for the hard questions,
when the correct match is less obvious for both the models and humans.
% \dd{Would suggest you to pick some numbers from the figure and talk about them here. These are key results of our study and should be discussed more in the text.} 
This further suggests that for the best results in practice, 
it may be useful to consider 
first identifying the difficulty of the question
and then determine if additional information is necessary.

% \vccomment{weird to start a paragraph by referencing a figure.. (same with next paragraph). I'm more generally confused by what the point of this paragraph is..}
% Figure~\ref{fig:accuracy} visualizes the mean accuracy results along with the baseline performance for each type of question (dotted horizontal lines), where the baseline is simply the accuracy of selecting the candidate article with the highest affinity score (i.e., the accuracy of the matching model). 
% Note that the model is only accurate around 33.3\% of the time in hard questions (red dotted line) while being accurate for all easy questions (blue dotted line). 
% This indicates that providing additional information with Syntactic or Semantic methods can be particularly helpful for hard questions -- cases when the correct match is less obvious. 
% On the other hand, it is more efficient to rely on the matching model
% for easy questions.
% This suggests that for the best results in practice, 
% it may be useful to first identify the difficulty of the question
% and then determine if additional information is necessary.

\subsubsection{Time Difference}
\label{sec:time}

We record the average response time (in seconds/question) for participants in each treatment group. 
We observe that on average the participants using SHAP, Co-occurrence, and Semantic methods
respond more quickly
compared to the control group for both types of questions (Figure~\ref{fig:time}).
% \footnote{Adhering to the details of the preregistration, we only report the statistics and do not conduct a test. No statistical tests are conducted on this data, due to a much larger sample size required to guarantee\ns{not sure this is convincing.. did we mention tests for time in the prereg? if not, just say that following the preregistration, we only report blah and do not conduct a test.  actually thinnking a bit more, maybe just omit this footnote as it seems a distraction rather than information} a desirable level of statistical power.} 
For both easy and hard questions,
the participants using the Semantic method
take the shortest average time (26.6 seconds for easy and 32.9 seconds for hard),
which is approximately a 20\% improvement over the control (34.4 seconds for easy and 41.7 seconds for hard).
The participants using BERTSum
take the longest (36 seconds for easy and 48.9 seconds for hard), 
where they experience a 17\% increase in time for the hard questions. 
% Note that the participants using the Semantic method take the shortest average time
% while the participants using BERTSum take the longest. 
% \dd{Same suggestion as above: talk about the raw numbers from Figure 6b here. People reading this should know, for instance, is the time saved of the order 2-5\%, or 20-30\%?}
Given that the Semantic method is also able to 
significantly boost the task accuracy, 
it is the most effective method 
among the tested ones. 
On the other hand, as BERTSum 
simultaneously decreases the task accuracy 
and increases the completion time,
it may be considered the least effective method.
% A potential reason may be 
% because the highlights failed to pick useful information, 
% users were forced to spend extra time processing both the highlights and the text.
% \vccomment{I would hesitate to follow with this because we cannot verify this is what actually happened to users.. I would move this to discussion if you wanted to keep it}This emphasizes being both correct and succinct about picking relevant information to present to the users. 
% Otherwise, they can either be inaccurate by relying on less useful information,
% or be slower by being forced to process more information.

% \begin{figure}[t]
%     \centering
%     \includegraphics[width=0.8\columnwidth]{figs/time.pdf}
%     \caption{Average time spent on each question for different treatment groups. We observe a decrease in average time for SHAP, Syntactic, and Semantic compared to control in both types of questions.}
%     \label{fig:time}
% \end{figure}

\subsubsection{Qualitative Responses}
\label{sec:qualresponse}

\begin{figure}[h]
    \centering
    \includegraphics[width=0.80\textwidth]{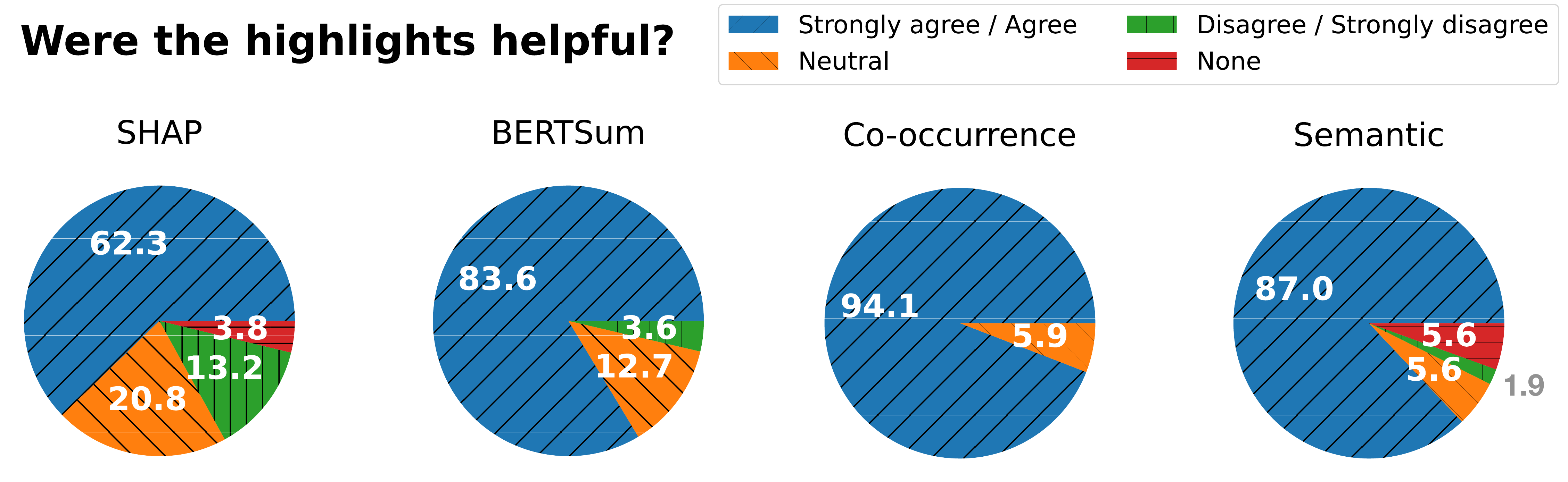}
    \caption{\small Participants' responses (in percentage) for ``Were the highlights helpful?''. For all the methods, the majority of the participants respond positively to the question regardless of their actual task accuracy.}
    \label{fig:qual1}
\end{figure}

At the end of the user study, the participants are asked several qualitative questions about the task. 

\textit{``Were the highlights helpful?''} Participants from all of the treatment groups generally respond positively to this question -- 
Figure~\ref{fig:qual1} shows the proportion of different responses from the participants in each group, 
and positive responses in blue form the majority in all groups. 
While the participants \emph{believe} the highlights to be helpful, 
their task performance shows the contrary for participants using SHAP and BERTSum. 
Such discrepancy between the subjective perception of a tool's utility 
and the objective utility measured by task-grounded performance metrics
corroborate similar previous observations on different assistive tools~\citep{kaur2020interpreting, bansal2021does}.
% \dd{We can talk about how this corroborates some of the past findings on explanations.}
% important to note for future experimental design.
% One potential reason for such discrepancy is the overreliance on the tool, 
% which has been observed in other similar studies involving participants utilizing automatically generated information to perform tasks~\citep{kaur2020interpreting, li2021algorithmic, bansal2021does} \vccomment{i dont think we have evidence from our study to back this up and dont think it adds anything}.
% We recommend future studies to develop and establish objective metrics about the helpfulness of assistive information (or other decision-support tools) for a particular need, 
% rather than solely relying on the participants' subjective perceptions. \vccomment{i think we can get rid of this sentence because it's not new.}
% \vccomment{i made direct edits to this para, please make sure it is accurate}
% See Appendix~\ref{appdx:results} for statistics on other questions 
% asked to the participants.

\begin{figure}[h]
    \centering
    \includegraphics[width=0.9\textwidth]{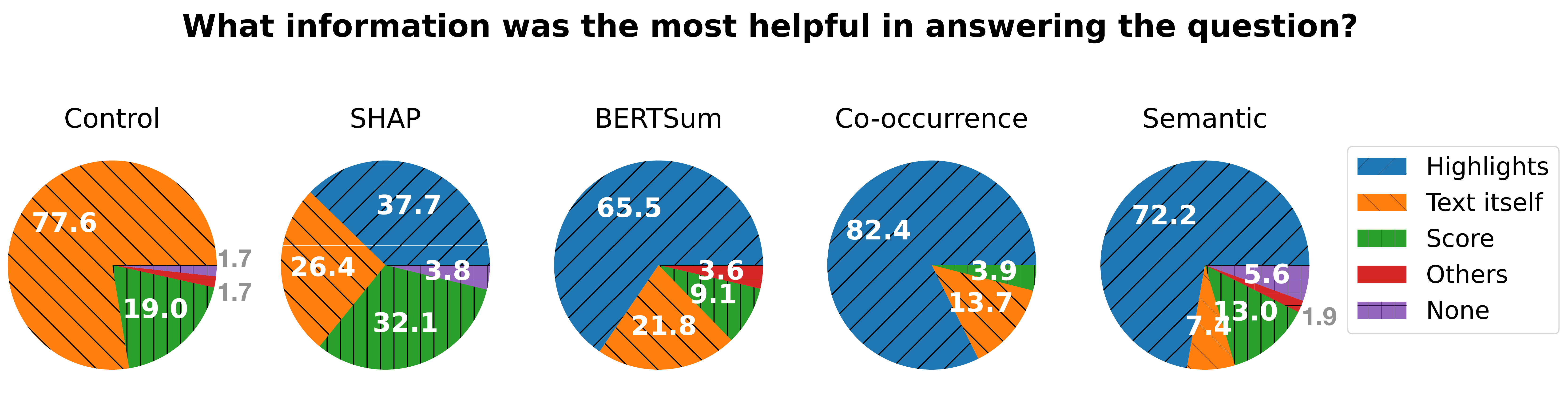}
    \caption{\small Participants' responses (in percentage) to a question ``What information was the most helpful?''}
    \label{fig:quals2}
\end{figure}

\noindent \textit{``What information was the most helpful in answering the question?''} 
While the majority of the participants using BERTSum, Co-occurrence, and Semantic methods 
respond that the highlights were the most helpful,
the participants using SHAP have more diverse responses that showed no particular
preference (Figure~\ref{fig:quals2}).
It is interesting to note that the participants using either Co-occurrence or Semantic
methods find the role of highlights to be significantly helpful when compared to other methods.
% (these methods
% are also the most effective in terms of task performance).

\begin{figure}[h]
    \centering
    \includegraphics[width=0.80\textwidth]{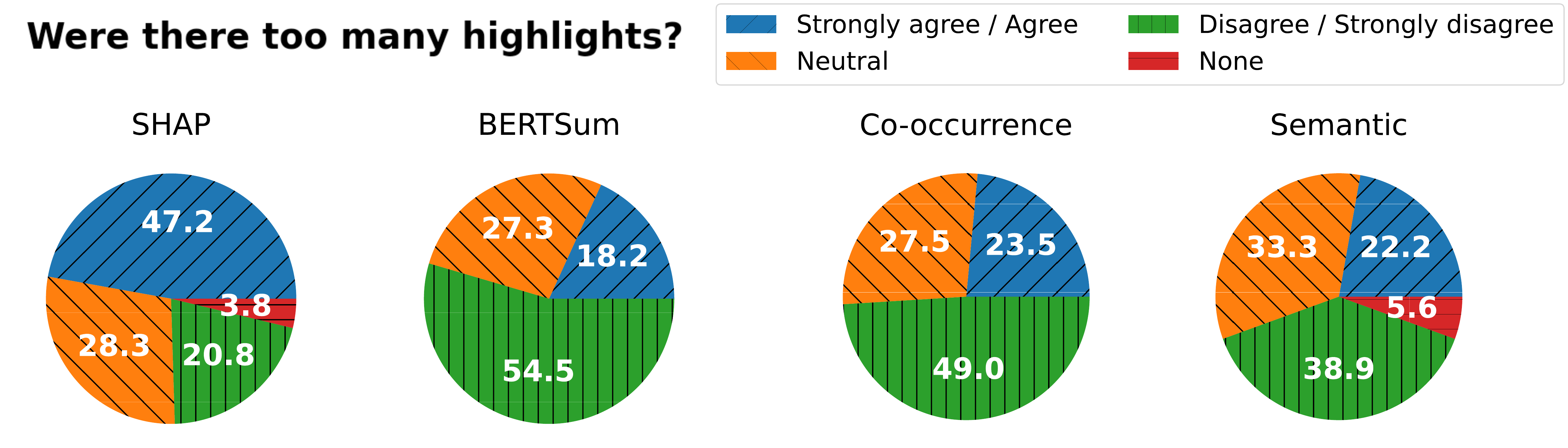}
    \caption{\small Participants' responses (in percentage) to a question ``Were there too many highlights?''}
    \label{fig:quals3}
\end{figure}

\noindent \textit{``Were there too many highlights?''}
We find that the participants using SHAP most strongly agree to this sentiment (Figure~\ref{fig:quals3}).
One factor that could have contributed to this is 
the default output values from SHAP used to generate the highlights, 
which are not post-processed for more
succinct representation of information.
Appropriate post-processing of the attribution scores 
may be necessary
to better account for this---the impact of the amount of 
highlights on the task performance is an open research question that requires future work.

%%% Discussion %%%
\section{Conclusion}
\label{sec:conclusion}

Motivated by 
practical concerns in document matching tasks with human decision makers,
% the lack of assistance
% \ns{this seems to be an extremely strong statement} \jkcomment{fixed}
% offered by document-matching models,
we conducted a user study to
investigate the utility of 
different kinds of assistive information
for the summary-article matching task. 
% designed the summary-article matching task where several assistive information 
% can be evaluated for their helpfulness on real users at scale.
We found that even well-established black-box model explanations
can potentially impair the users' decisions,
while task-specific approaches can 
effectively assist them. 
Existing methods are typically not explicitly 
optimized for the task's objective:
Model explanations are contingent on the matching model;
it attempts to explain what the \emph{model} considers important,
not necessarily what \emph{human} users find important
to perform well in the task. 
General text summarization methods can be helpful for succinctly expressing 
a high-level topic of the article,
but may lack the precision of picking the details directly related to the given summary.
Furthermore, we observed that
the users' subjective perception on the utility of (assistive) information
was misaligned with their performance on the task.
% These results altogether 
% emphasize the need to 
% carefully examine the application of black-box explanations
% for downstream use cases. 
These results altogether emphasize that 
it is important for the developers 
of such assistive tools 
to articulate the specific use (and users) it serves, 
and rigorously evaluate their proposals. 

We believe 
that the summary-article matching task
can be used as a first-pass test 
to validate
promising methods (and promote development of new approaches). %for real decision makers and systems in practice.
% useful in more realistic applications like peer review. 
Relaxing some assumptions in our setup 
can provide further insights on strengths and weaknesses of 
individual methods in more complex scenarios (e.g., allowing 
multiple or no ground-truths in questions). 
Complex and lengthy documents in practice (e.g., 
academic papers and reviewer profiles) may require
longer data collection phase and 
more scalable adaptations of the presented methods.
For instance, visualizing all highlights with different colors at once 
(as we did in the summary-article matching task)
may not be practical for longer documents.
Necessary changes in the user interface can address such issues, 
e.g., the highlights 
can be shown interactively
based on what information the user is interested in.
\section*{Acknowledgments}
We thank Kasun Amarasinghe, Siddhant Arora, Keegan Harris, Nari Johnson, and Yilun Zhou 
for helpful feedback and discussions.
This research was approved by CMU Institutional Review Board (IRB). 
%This work was supported by NSF grant 1942124. 
% \ns{Ameet's grants?}
This work was supported in part by the National Science Foundation grants CIF1942124, IIS1705121, IIS1838017, IIS2046613, IIS2112471, CIF1763734 and funding from Meta, Morgan Stanley, Amazon, and Google. Any opinions, findings and conclusions or recommendations expressed in this material are those of the author(s) and do not necessarily reflect the views of any of these funding agencies.

%Bibliography
\bibliographystyle{tmlr}  
\bibliography{anthology, custom}  

\newpage
\appendix
\noindent {\Large \bf Appendices}

% %%% Appendix %%%
\section{Proxy Test of Black-box Model Explanations}
\label{appdx:posthoc}
%% How we decided to use SHAP for post hoc
There are several black-box model explanations to consider for the task. 
While testing all of them on real users can be an interesting research on its own, 
as we are more broadly interested in what distinct types of information could be helpful, 
we decide to select one representative method in the literature.
As there is no absolute answer to which method is superior,  
we conduct a simple proxy test of what method can be a better choice for the task.

\begin{figure}[h]
    \centering
    \includegraphics[width=0.4\columnwidth]{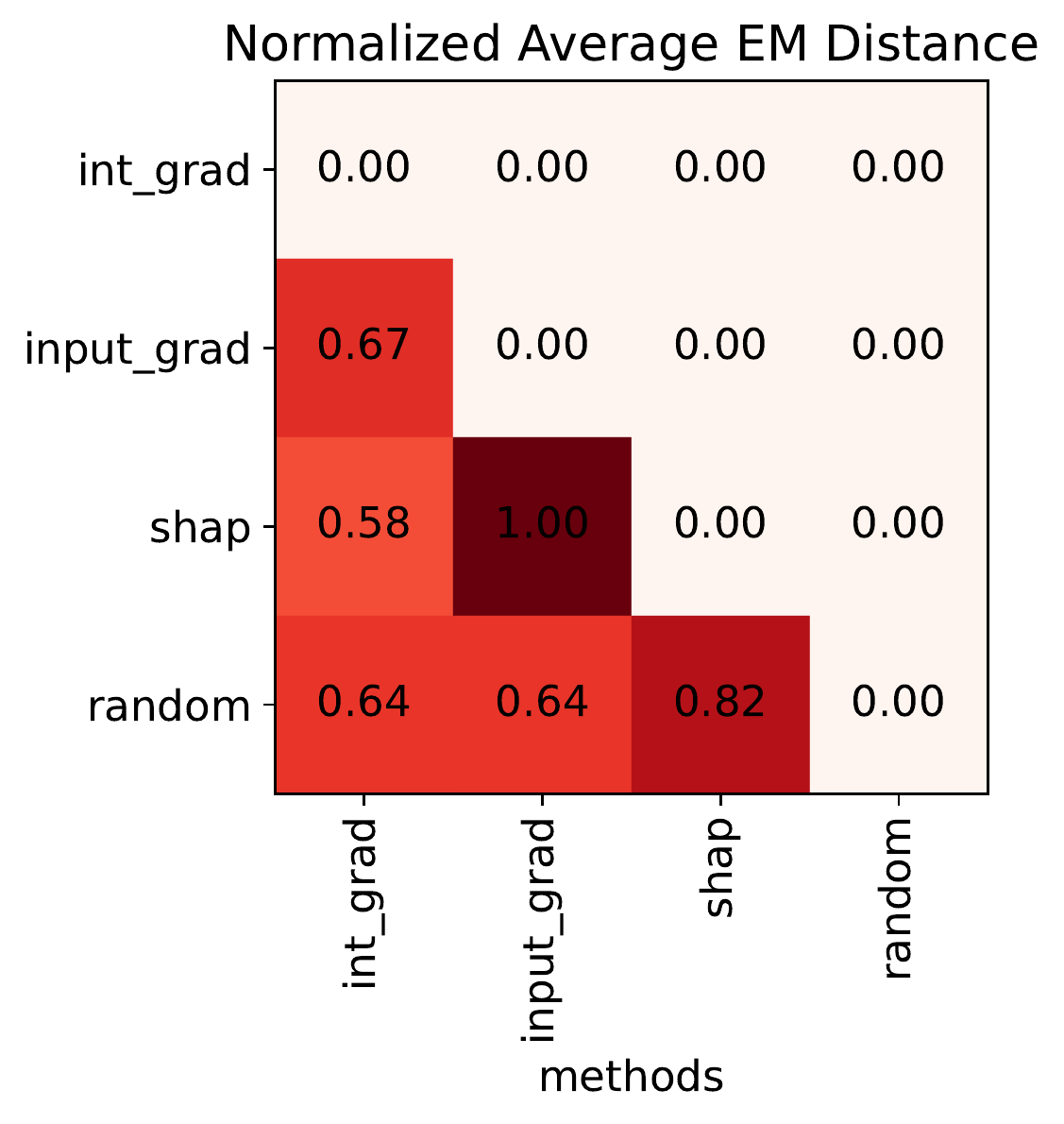}
    \caption{Proxy quality test for the black-box model explanations using average EM distances between the distributions of attribution scores of input tokens.
    The higher the values (the darker the color), the more different the distribution of the attribution scores. 
    SHAP shows the most distinct distribution from the random attributions (bottom row).}
    \label{fig:appdx_posthoc}
\end{figure}

We consider the following feature attribution methods: Integrated Gradients~\citep{sundararajan2017axiomatic}, Input x Gradients~\citep{shrikumar2016not}, and SHAP~\citep{lundberg2017unified}. 
In Figure~\ref{fig:appdx_posthoc}, we plot the mean EM distance (averaged across 50 different random attributions, normalized to be between 0 and 1)
between the distribution of attribution scores for the input tokens 
in our ground-truth articles. 
The higher the value (darker the color), the more distinct the distribution of 
the attribution scores computed by respective methods.
Notice that SHAP shows the most distinct distribution from random attributions
compared to other methods, indicating it may be a better choice
that carry more information about the important tokens.
We have also qualitatively verified that the highlights from other two methods were
not as meaningful as SHAP on the articles.

Note also that SHAP is a promising candidate to apply to the task
due to its popularity and its common presence in more sophisticated domains like
biology, physics, chemistry, and finance~\citep{jesus2021can, novakovsky2022obtaining, yang2022unbox, zablocki2022explainability, pucci2022artificial}.

% \subsection{Method Examples}
% \label{appdx:method-examples}
% %% UI examples for different methods
%%%%%%%

\newpage 
\section{Method Examples}
\label{appdx:method-examples}

We show below some example highlights presented to the users using different methods.

\begin{figure}[h]
    \centering
    \includegraphics[width=\textwidth]{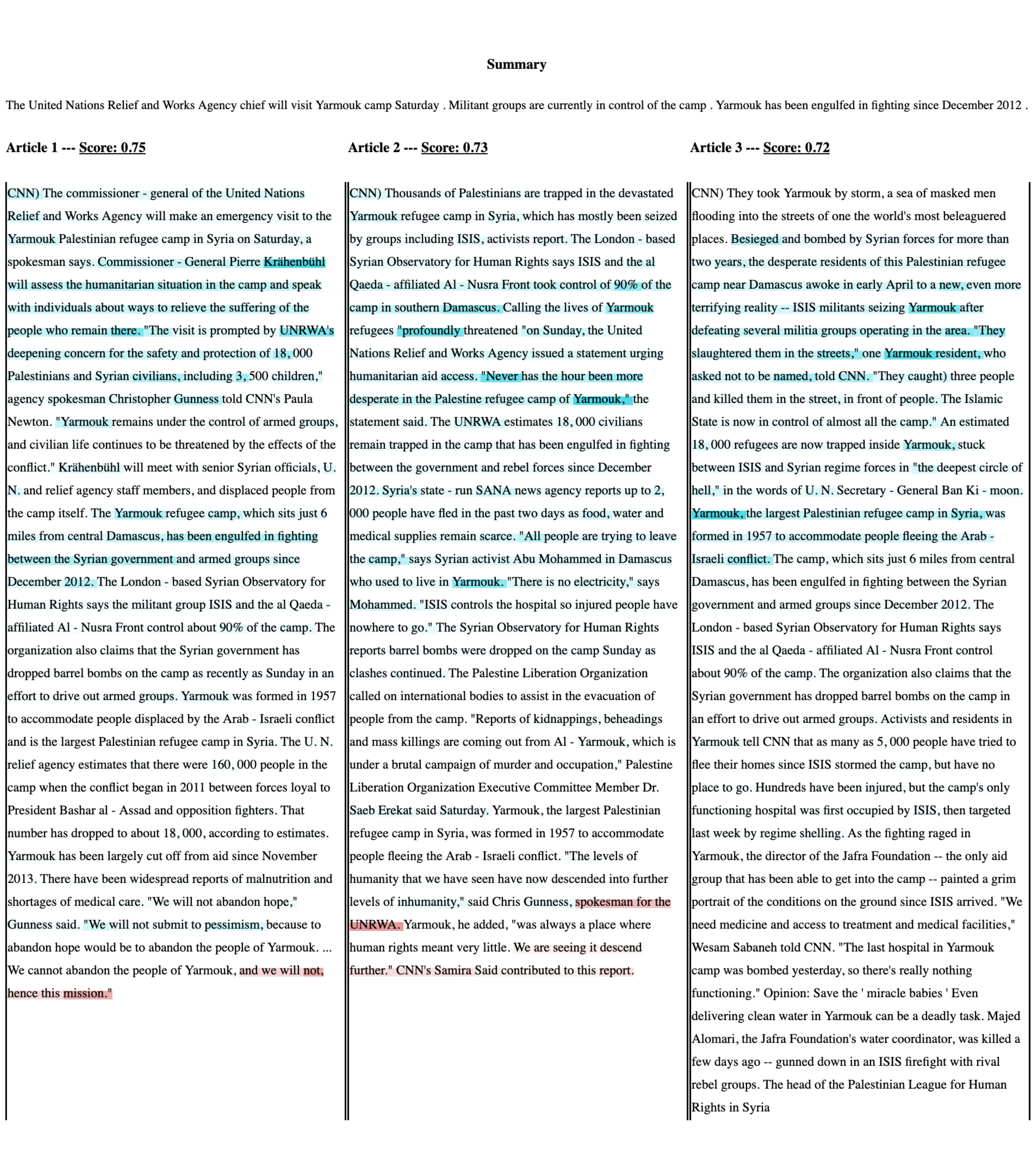}
    \caption{Example highlights for SHAP.}
    \label{fig:235shap}
\end{figure}

\newpage
\begin{figure}[h]
    \centering
    \includegraphics[width=\textwidth]{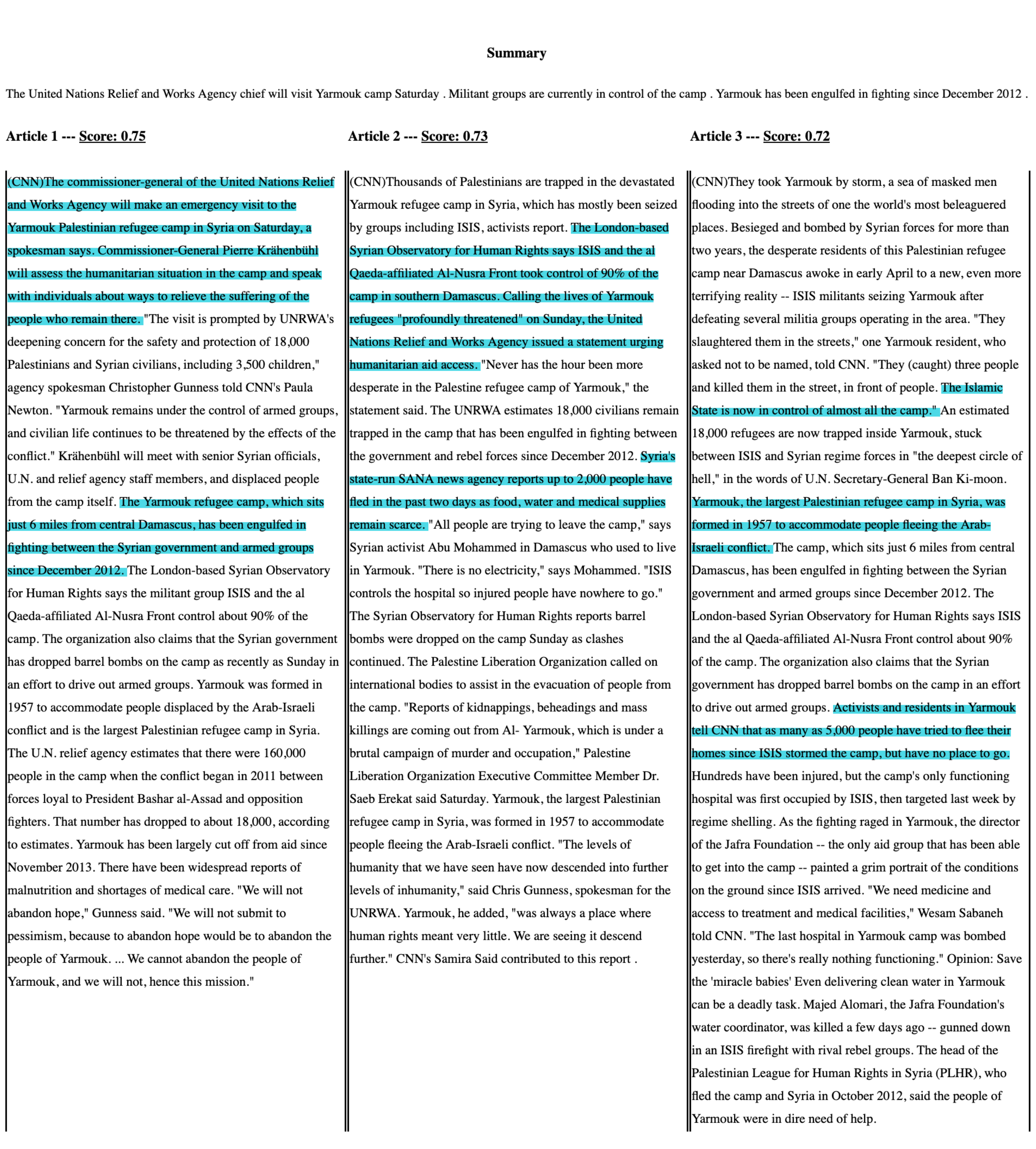}
    \caption{Example highlights for BERTSum.}
    \label{fig:235bertsum}
\end{figure}

\newpage
\begin{figure}[h]
    \centering
    \includegraphics[width=\textwidth]{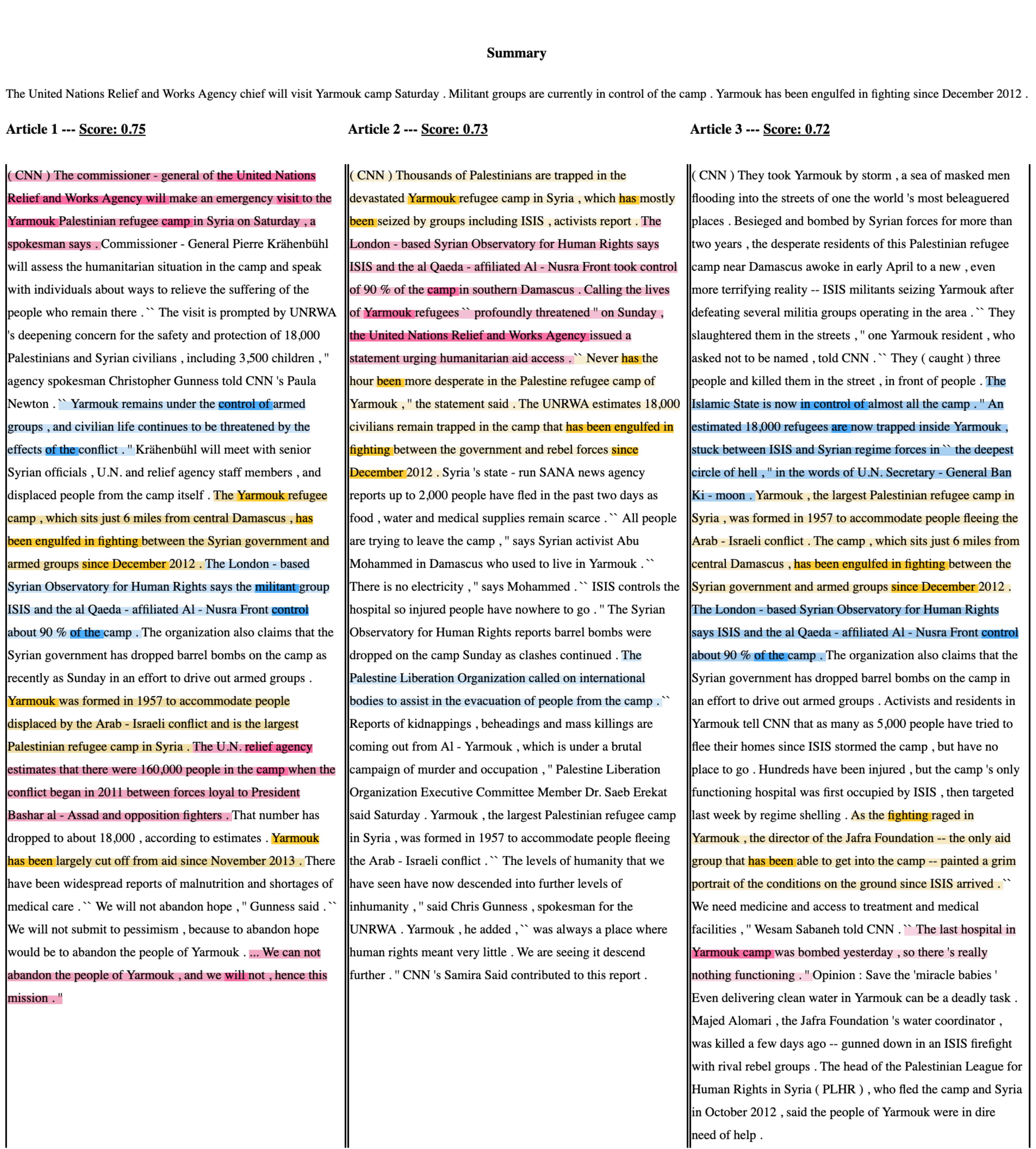}
    \caption{Example highlights for Co-occurrence method.}
    \label{fig:235syntactic}
\end{figure}

\newpage
\begin{figure}[h]
    \centering
    \includegraphics[width=\textwidth]{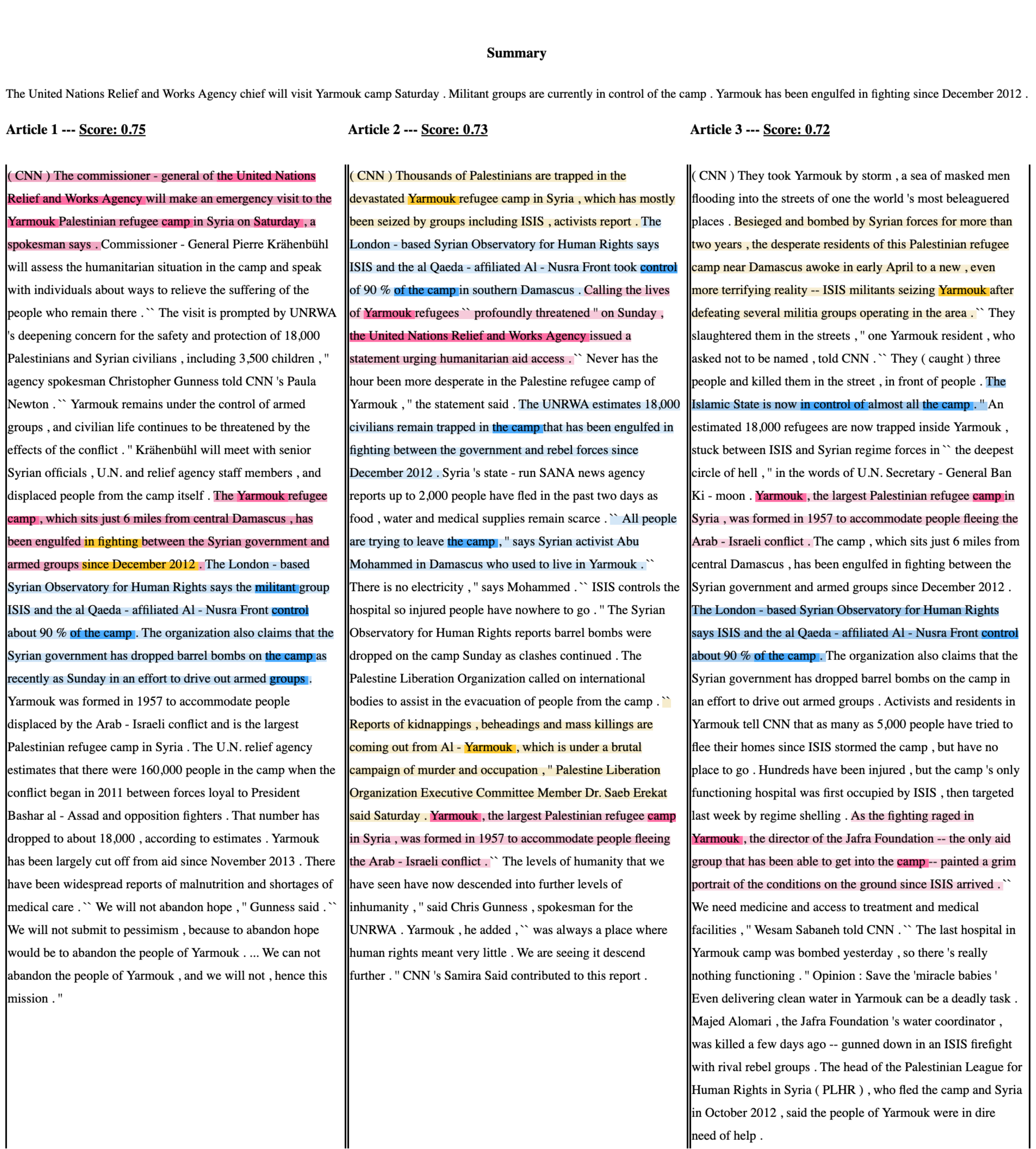}
    \caption{Example highlights for Semantic method.}
    \label{fig:235semantic}
\end{figure}
%%%%%%%%

\newpage

\section{User Study Details}
\label{appdx:study-design}
%% More details on the user study (like bonus design, payment, power analysis etc.)

\subsection{Pilots and Sample Size}
\label{appdx:samplesize}

\begin{figure}[h]
    \centering
    \includegraphics[width=0.6\columnwidth]{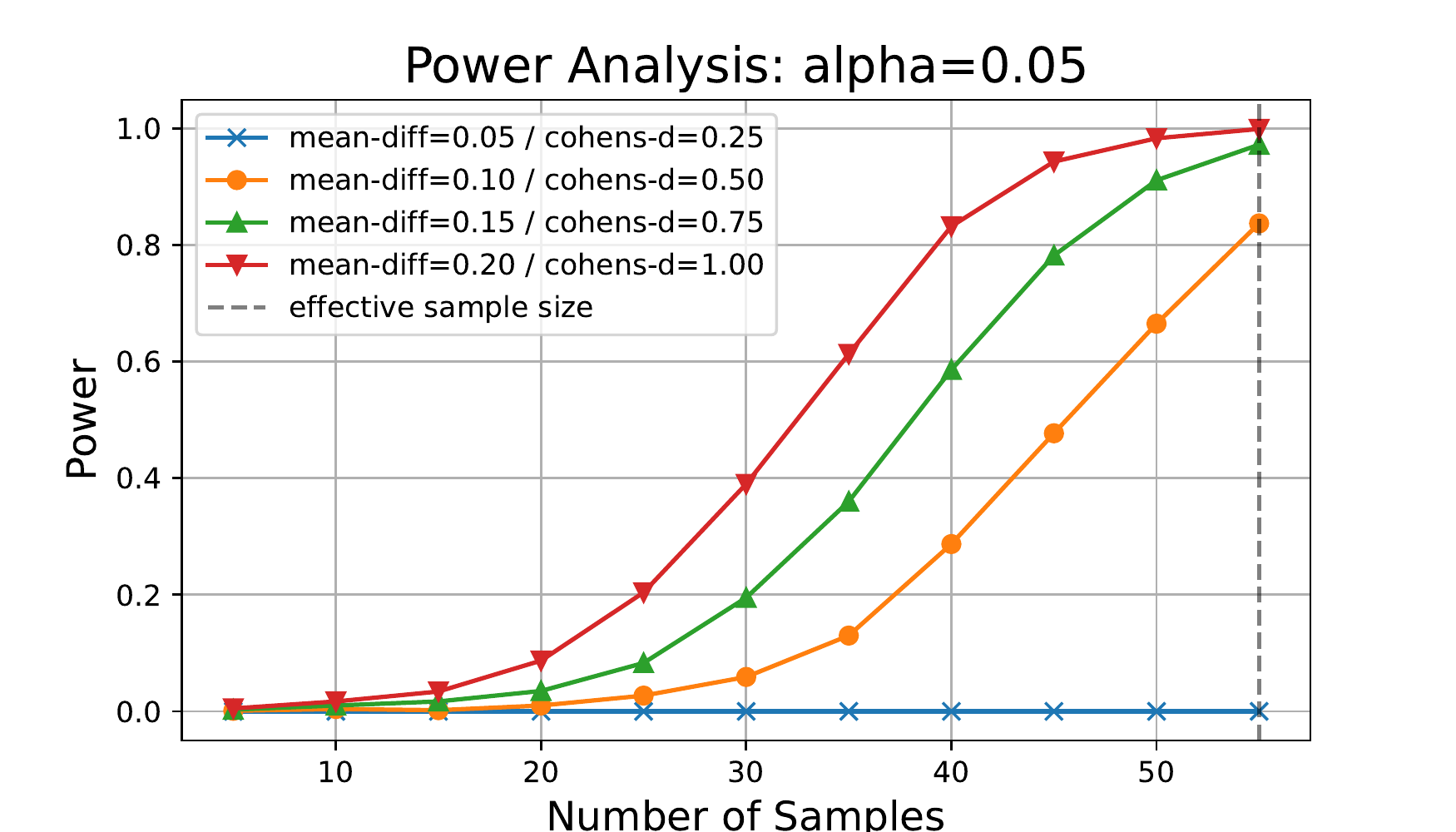}
    \caption{Power analysis for the effective sample size. We collect 55 samples per group (vertical dotted line) for a statistical power over 0.8 for the effect size (Cohen's $d$) of 0.5 (orange line with circle markers).}
    \label{fig:power}
\end{figure}

Prior to conducting the actual user study, 
we ran pilot studies on a smaller number of participants.
Using the data points collected from these studies, we conducted a Monte Carlo simulation-based power analysis to determine the effective sample size.  
We determined to recruit 55 participants per condition (so total of 275 = 55 $\times$ 5 conditions) for a statistical power over 0.8 with the effect size (Cohen's d) of 0.5  (orange line with circle markers in Figure~\ref{fig:power}). 
This effect size corresponds to 0.1 difference in the mean accuracy between the control and the treatment. 

\subsection{Demographic Background}
\label{appdx:demoinfo}

\begin{figure}[h]
    \centering
    \includegraphics[width=0.48\textwidth]{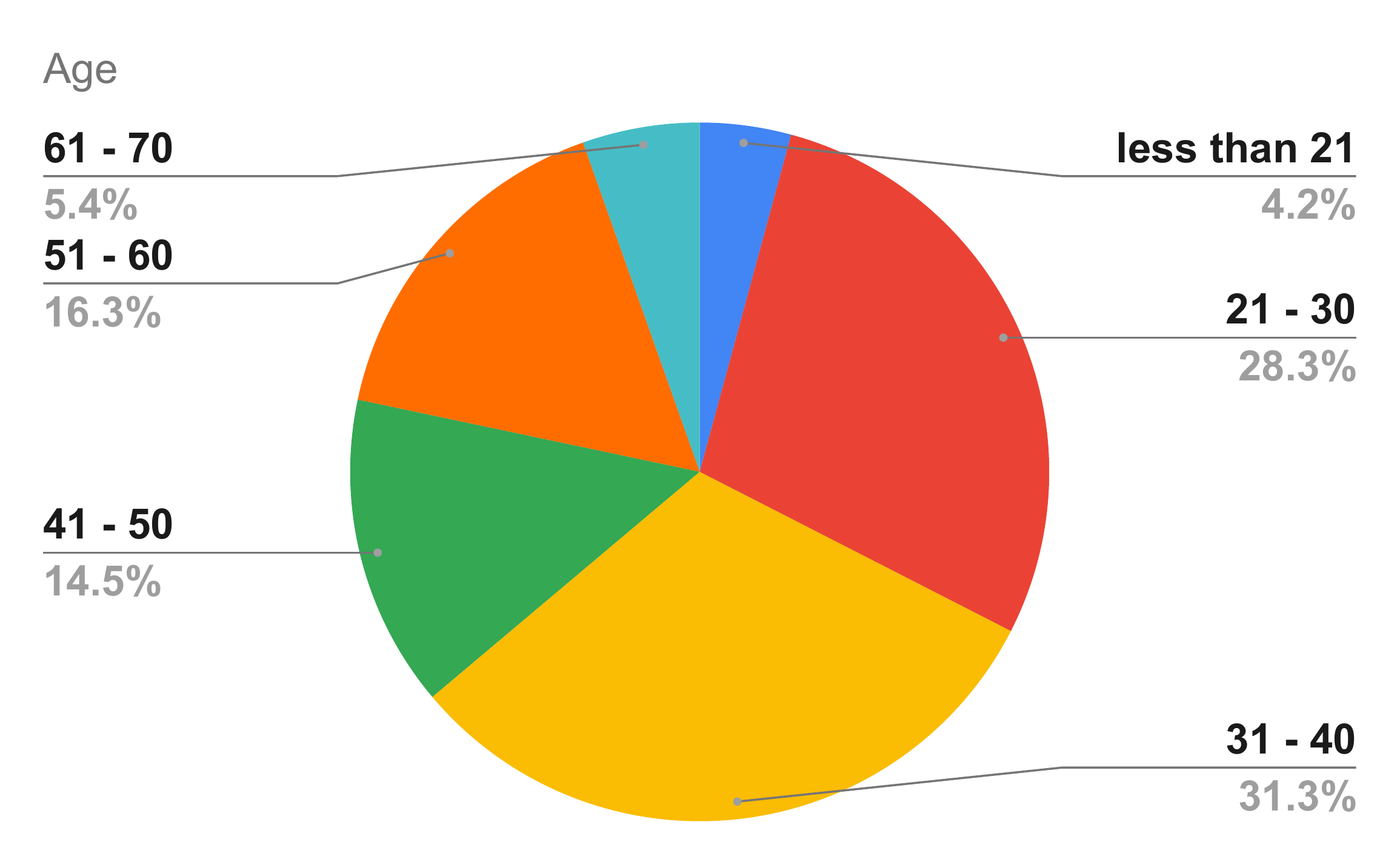}
    \includegraphics[width=0.48\textwidth]{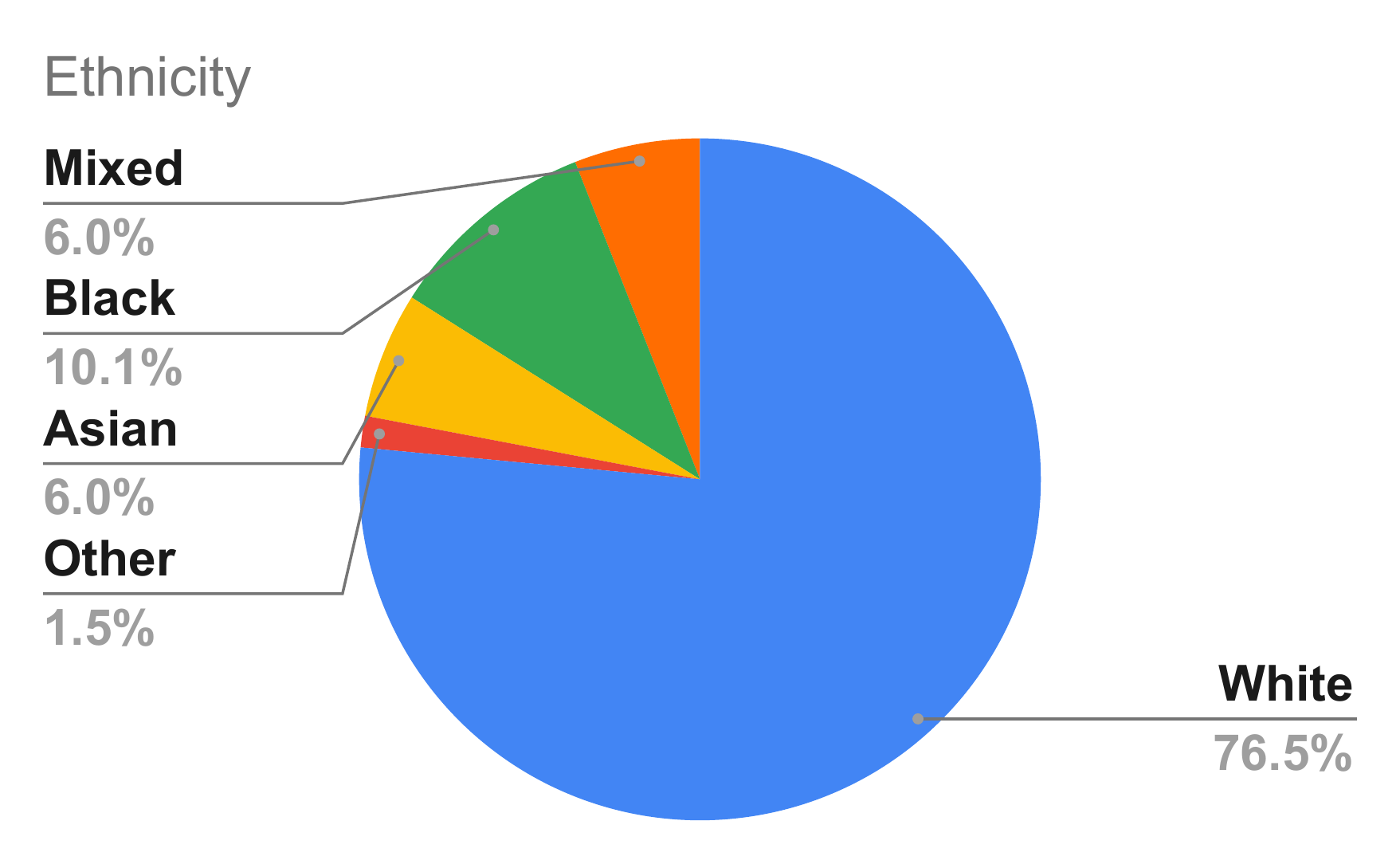}
    \includegraphics[width=0.48\textwidth]{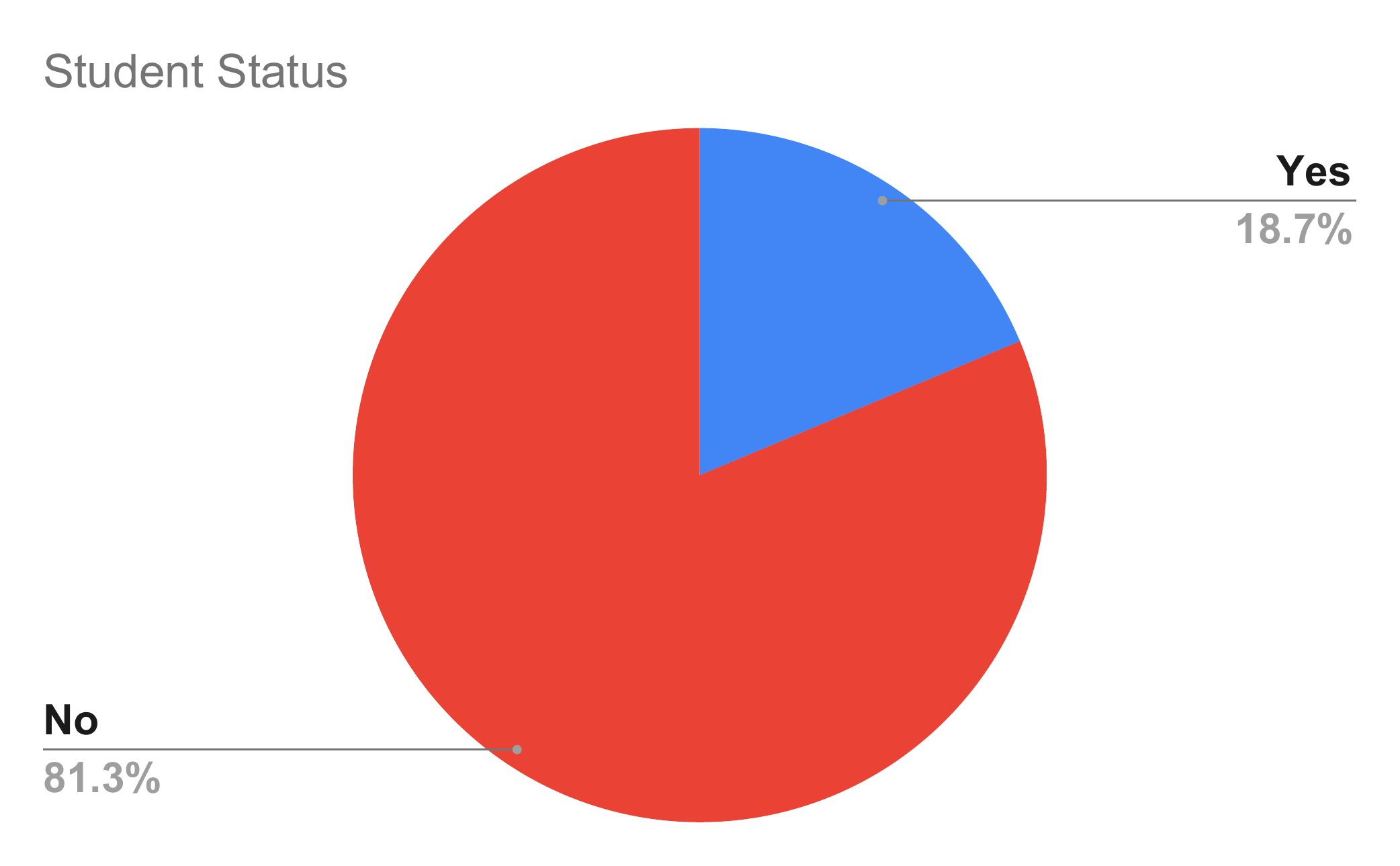}
    \includegraphics[width=0.48\textwidth]{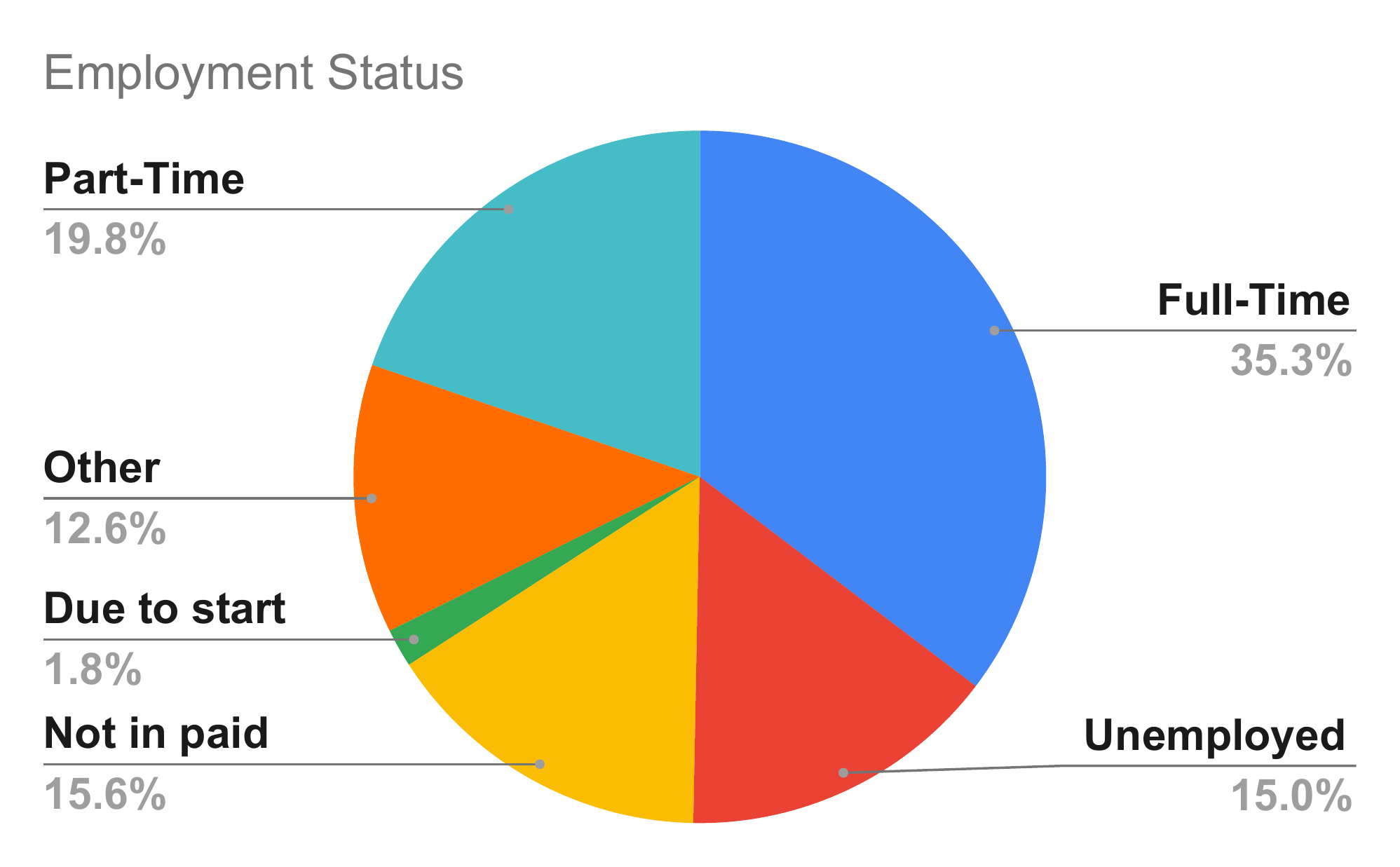}
    \caption{Demographic background of the participants (age, ethnicity, student status, and employment status).}
    \label{fig:demoinfo}
\end{figure}

In Figure~\ref{fig:demoinfo}, we provide demographic background of the participants (age, ethnicity, student status, employment status) recruited for the study.
$275$ participants were recruited from a balanced pool of adult males and females located in the U.S. with minimum approval ratings of $90\%$ using Prolific (\url{www.prolific.co}).

\subsection{Tutorial}

We provide the participants with a set of instructions 
laying out what the highlights indicate and how one might use them for the task. 
The instruction is followed by two sample questions on which the participants could
take unlimited time to get an understanding of what the questions look like. 
For the sample questions, the participants were provided the correct answers
and the justification behind them as feedback.

\subsection{Payments}
\label{appdx:payments}

Base payment per participants was \$3.15, determined based on
the minimum hourly payment set by the platform and the median completion time of all participants, resulting in an average reward of \$12.07 per hour.
To encourage quicker and more accurate responses, we designed bonus payments so that
each participant could earn additional \$ (base payment for the question $\times$ multiplier) for each correctly answered questions, where the multiplier is determined by the response time on the question (Table~\ref{tab:bonus}). 
One could ideally earn up to $\times 1.5$ the base payment by answering all questions correctly,
all within 30 seconds. 
All payments (base and bonus) were processed after the data collection was complete, accounting for invalid responses.

\begin{table}[h]
    \centering
    \begin{tabular}{c c c c c c }
        Response Time (seconds) & < 30 & < 60 & < 90 & < 120 & > 120 \\ \hline \hline
        Multiplier &  x0.5 & x0.4 & x0.3 & x0.2 & x0.0
    \end{tabular}
    \caption{Reward multiplier based on response time for correct answers. Incorrect answers have the multiplier of zero.}
    \label{tab:bonus}
\end{table}

% \subsubsection*{Acknowledgments}
% Use unnumbered third level headings for the acknowledgments. All
% acknowledgments, including those to funding agencies, go at the end of the paper.
% Only add this information once your submission is accepted and deanonymized. 

% \bibliography{main}
% \bibliographystyle{tmlr}

% \appendix
% \section{Appendix}
% You may include other additional sections here.

\end{document}